\documentclass[sigconf]{acmart}
\AtBeginDocument{%
  \providecommand\BibTeX{{%
    \normalfont B\kern-0.5em{\scshape i\kern-0.25em b}\kern-0.8em\TeX}}}

\copyrightyear{2023}
\acmYear{2023}
\setcopyright{acmlicensed}
\acmConference[CIKM '23] {Proceedings of the 32nd ACM International Conference on Information and Knowledge Management}{October 21--25, 2023}{Birmingham, United Kingdom.}
\acmBooktitle{Proceedings of the 32nd ACM International Conference on Information and Knowledge Management (CIKM '23), October 21--25, 2023, Birmingham, United Kingdom}
\acmPrice{15.00}
\acmISBN{979-8-4007-0124-5/23/10}
\acmDOI{10.1145/3583780.3615040}

\usepackage{subcaption}
\usepackage{enumitem}
\usepackage{algorithm}
\usepackage{algpseudocode}

\usepackage{multirow}
\usepackage{bbm}
\usepackage{amsthm}
\usepackage{enumitem}
\usepackage{colortbl}

\definecolor{ccon}{HTML}{fee9d4} % orange
\definecolor{cood}{HTML}{d8f0d3} % green
\definecolor{cid}{HTML}{dae8f5} % blue
\definecolor{cdel}{rgb}{0.83, 0.32, 0.16} % darker orange
\definecolor{cadd}{rgb}{0, 0.47, 0.34} % darker green

\definecolor{orange}{HTML}{FF9500}

\newcommand{\cell}[1]{\begin{tabular}{@{}r@{}}#1\end{tabular}}

\newcommand{\TITLE}{RoCourseNet: Robust Training of a Prediction Aware Recourse Model}

\newcommand{\cf}{x^\text{cf}}

\begin{document}

\title{\TITLE}

\author{Hangzhi Guo}
\email{hangz@psu.edu}
\affiliation{%
  \institution{The Pennsylvania State University}
  \city{University Park}
  \state{PA}
  \country{USA}
}

\author{Feiran Jia}
\email{fzj5059@psu.edu}
\affiliation{%
  \institution{The Pennsylvania State University}
  \city{University Park}
  \state{PA}
  \country{USA}
}

\author{Jinghui Chen}
\email{jzc5917@psu.edu}
\affiliation{%
  \institution{The Pennsylvania State University}
  \city{University Park}
  \state{PA}
  \country{USA}
}

\author{Anna Squicciarini}
\email{acs20@psu.edu}
\affiliation{%
  \institution{The Pennsylvania State University}
  \city{University Park}
  \state{PA}
  \country{USA}
}

\author{Amulya Yadav}
\email{amulya@psu.edu}
\affiliation{%
  \institution{The Pennsylvania State University}
  \city{University Park}
  \state{PA}
  \country{USA}
}

\renewcommand{\shortauthors}{Hangzhi Guo, Feiran Jia, Jinghui Chen, Anna Squicciarini, and Amulya Yadav}

\begin{abstract}
  %Counterfactual (CF) explanation is a preferable explanation method for end-users which explains the predictions of machine learning (ML) models by providing a recourse case to individuals who are adversely impacted by predicted outcomes. 
%Counterfactual (CF) explanation is a preferable explanation method for end-users which explains the predictions of machine learning (ML) models by providing a recourse case to individuals who are adversely impacted by predicted outcomes. 
Counterfactual (CF) explanations for machine learning (ML) models are preferred by end-users, as they explain the predictions of ML models by providing a recourse (or contrastive) case to individuals who are adversely impacted by predicted outcomes. 
%Existing CF explanation methods generate recourses under the assumption that the underlying target ML model remains stationary over time. However, due to commonly occurring distributional shifts in training data, ML models constantly get updated in practice, which might render previously generated recourses invalid, and in turn, diminish end-users trust in our algorithmic framework.
Existing CF explanation methods generate recourses under the assumption that the underlying target ML model remains stationary over time. However, due to commonly occurring distributional shifts in training data, ML models constantly get updated in practice, which might render previously generated recourses invalid and diminish end-users trust in our algorithmic framework. 
%To address this problem, we propose RoCourseNet, an end-to-end training framework that optimizing for predictions and robust recourses to future data shifts. In particular, we propose three novel contributions: 
To address this problem, we propose RoCourseNet, a training framework that jointly optimizes predictions and recourses that are robust to future data shifts. 
% We have three main contributions:
This work contains four key contributions:
(1) We formulate the robust recourse generation problem as a tri-level optimization problem which consists of two sub-problems: (i) a bi-level problem that finds the worst-case adversarial shift in the training data, and (ii) an outer minimization problem to generate robust recourses against this worst-case shift.
(2) We leverage adversarial training to solve this tri-level optimization problem by: (i) proposing a novel \emph{virtual data shift (VDS)} algorithm to find worst-case shifted ML models via explicitly considering the worst-case data shift in the training dataset, and (ii) a block-wise coordinate descent procedure to optimize for prediction and corresponding robust recourses.
(3) We evaluate RoCourseNet's performance on three real-world datasets, and show that RoCourseNet consistently achieves more than 96\% robust validity and outperforms state-of-the-art baselines by at least 10\% in generating robust CF explanations.
(4) Finally, we generalize the RoCourseNet framework to accommodate any parametric post-hoc methods for improving robust validity.
% \footnote{We release the code through an anonymous repository in supporting the reproduction of this work (\url{https://www.dropbox.com/s/gsrpt55hf2ik7v4/RoCourseNet.zip?dl=0}).}

% (i) We formulate We formulate the robust recourse generation problem as a tri-level (min-max-min) optimization problem,
% which consists of two sub-problems: (i) a bi-level (max-min) problem which simulates a worst-case attacker to find an adversarially
% shifted model by explicitly simulating the worst-case data shift in the training dataset; and (ii) an outer minimization problem
% which simulates an ML model designer who wants to generate robust recourses against this worst-case bi-level attacker.

% (i) We propose a novel \emph{virtual data shift (VDS)} algorithm to find worst-case shifted ML models by explicitly considering the worst-case data shift in the training dataset.
% (ii) We leverage adversarial training to solve a novel tri-level optimization problem inside RoCourseNet, which simultaneously
% %optimize a tri-level optimization problem fr  
% generates predictions and corresponding robust recourses.
% %(iii) Finally, we evaluate RoCourseNet's performance on several real-world datasets, which shows that RoCourseNet outperforms state-of-the-art baselines by $\sim$10\% in generating robust CF explanations.
% (iii) Finally, we evaluate RoCourseNet's performance on three real-world datasets and show that RoCourseNet outperforms state-of-the-art baselines by $\sim$10\% in generating robust CF explanations.

\end{abstract}





% \received{2 February 2023}

\maketitle

\section{Introduction}
\label{sec:intro}

Existing work in Explainable Artificial Intelligence (XAI) has been focused on developing techniques to interpret decisions made by black-box machine learning (ML) models \citep{ribeiro2016lime, lundberg2017unified, koh2017understanding, kim2018interpretability}. In particular, counterfactual (CF) explanation methods find a new \emph{counterfactual} example $\cf$, which is similar to input instance $x$ but gets a different/opposite prediction from the ML model. 
Counterfactual explanation techniques \citep{wachter2017counterfactual, ustun2019actionable, mothilal2020explaining, karimi2021algorithmic} are often preferred by human end-users because of their ability to provide actionable recourse
\footnote{Note that counterfactual explanation \citep{wachter2017counterfactual} and algorithmic recourse \citep{ustun2019actionable} are closely related \citep{verma2020counterfactual, stepin2021survey}. Hence, we use these terms interchangeably.} 
to individuals who are negatively impacted by algorithm-mediated decisions. For example, CF explanation techniques can be used to provide algorithmic recourse for impoverished loan applicants who have been denied a loan by a bank's ML algorithm, etc.% or to provide recourse recommendations for school teachers working with students who are at the risk of dropping out from school, etc.

\begin{figure*}[ht!]
     \centering
     \begin{subfigure}[h]{0.32\textwidth}
         \centering
         \includegraphics[width=0.98\textwidth]{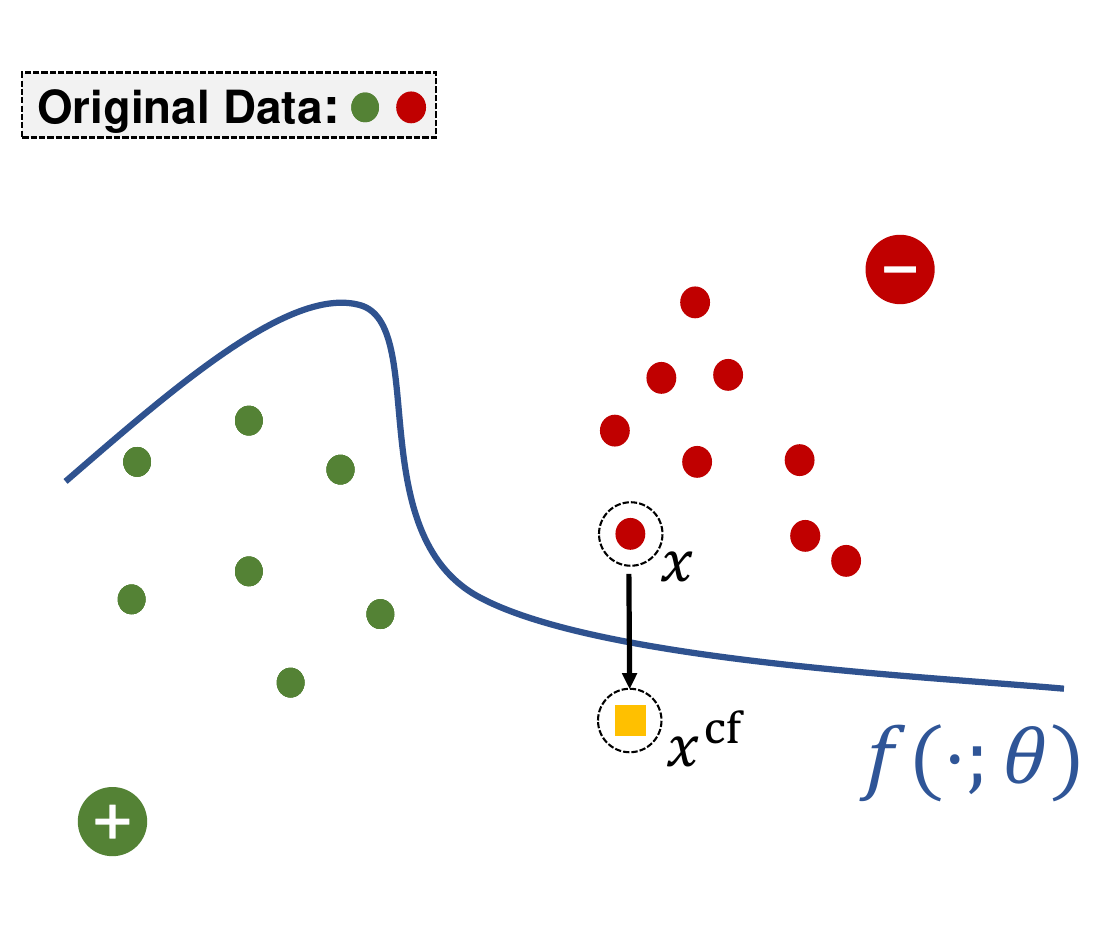}
         % \caption{The original data, decision boundary of model $f(\cdot,\theta)$, and a recourse $x^{\text{cf}}$ of the example $x$.}
         \caption{The decision boundary of model $f(\cdot,\theta)$ trained on the original data, and a recourse  $\cf$ of the example $x$.}
         \label{fig:original_dis}
     \end{subfigure}
     \hfill
     \begin{subfigure}[h]{0.32\textwidth}
         \centering
         \includegraphics[width=0.98\textwidth]{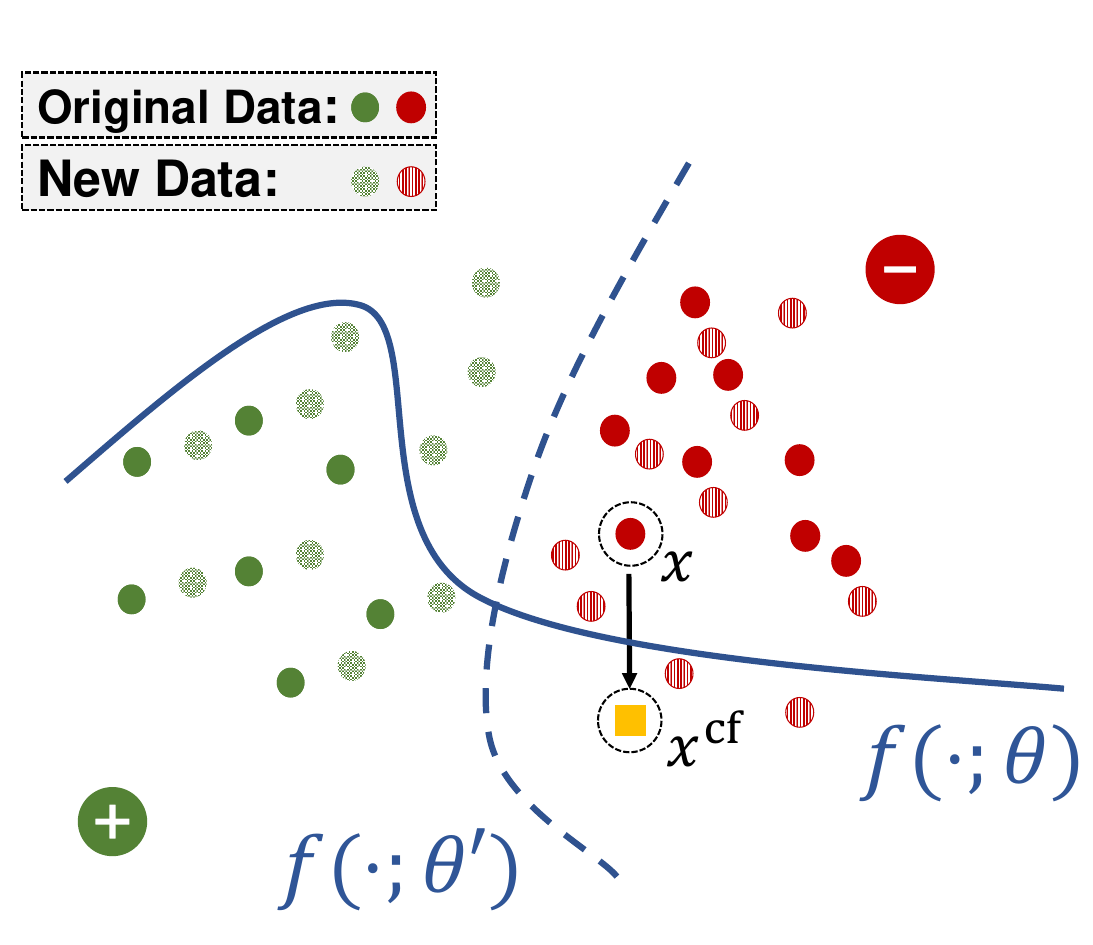}
         % \caption{The original/shifted data and the decision boundary of retrained new model $f(\cdot,\theta')$.}
         \caption{Updated decision boundary of retrained new model $f(\cdot,\theta')$ with newly available data (or a shifted data distribution).}
         \label{fig:new_dis}
     \end{subfigure}
     \hfill
         \begin{subfigure}[h]{0.32\textwidth}
         \centering
         \includegraphics[width=0.98\textwidth]{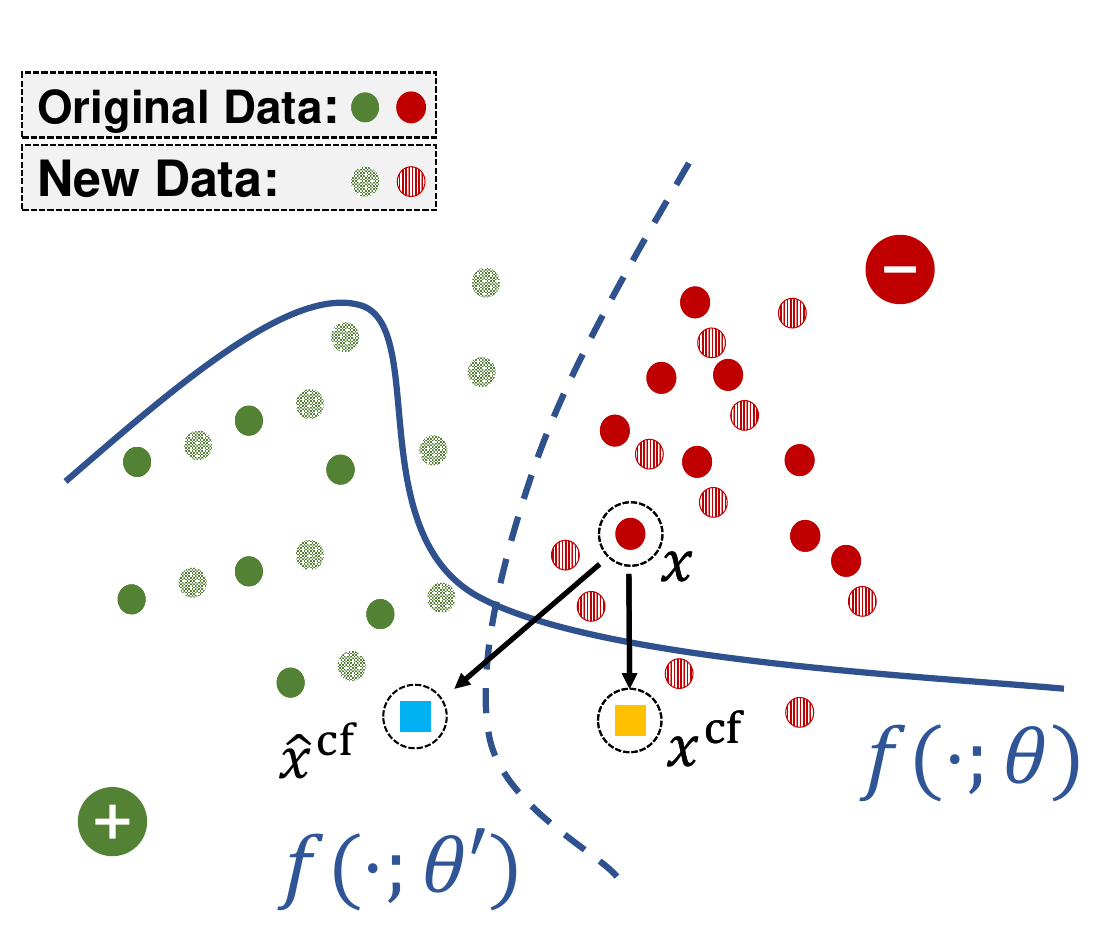}
         \caption{Under the shifts in data distribution and model, the recourse $\cf$ becomes invalid, but $\hat x^{\text{cf}}$ is valid. We call $\hat x^{\text{cf}}$ as a robust recourse.}
         \label{fig:rob_cf}
     \end{subfigure}
    \caption{Illustration of the robust recourse generation process.
    (a) Given an input data point $x$, CF explanation methods generate a new recourse $\cf$ which lies on the opposite side of decision boundary $f(.;\theta)$. 
    (b) As \emph{new data} is made available, the ML model's decision boundary is updated as $f(.;\theta')$. This shifted decision boundary $f(.;\theta')$ invalidates the chosen recourse $\cf$ (as $x$ and $\cf$ lie on the same side of the shifted model $f(.;\theta')$). (c) However, robust CF explanation methods generate a \emph{robust} recourse $\hat{x}^\text{cf}$ for input $x$ by anticipating the future shifted model $f(.;\theta')$.}
    \label{fig:datashiftissue}
\end{figure*}

Most CF explanation techniques assume that the underlying ML model is stationary and does not change over time \citep{barocas2020hidden}. However, in practice, ML models are often updated regularly when new data is available to improve predictive accuracy on the new shifted data distribution. This shifted ML model might render previously recommended recourses ineffective \citep{rawal2020can}, and in turn, diminish end users' trust towards our system. For example, when providing a recourse to a loan applicant who was denied a loan by the bank's ML algorithm, it is critical to ensure that the bank can honor that recourse and 
approve re-applications that fully follow recourse recommendations, even if the bank updates their ML model in the meantime. This necessitates the development of robust algorithms that can generate recourses that remain effective (or valid) for an end-user in the face of ML models being frequently updated. 
Figure~\ref{fig:datashiftissue} illustrates this challenge of generating robust recourses. \\
% \footnote{We further elucidate the motivations of robust recourse explanation through an end-user use-case scenario in Appendix~\ref{sec:app-motivation}.} \\

% Figure \ref{fig:datashiftissue} illustrates this challenge of generating robust recourses. Let $f(.;\theta)$ denote an ML model's decision boundary. Given an input data point $x$, CF explanation methods generate 
% a new counterfactual $\cf$ 
% (shown in \textcolor{orange}{yellow}) 
% as a recourse, which lies on the opposite side of decision boundary $f(.;\theta)$. However, as \emph{new data} is made available, the ML model's decision boundary may get shifted (denoted as $f(.;\theta')$). 
% This shifted decision boundary invalidates the chosen recourse $\cf$ (as $x$ and $\cf$ lie on the same side of $f(.;\theta')$). This is undesirable to an end-user because following recourse recommendations (as per $\cf$) would not lead to a favorable decision by the shifted ML model.
% Robust CF explanation methods address this problem by generating a \emph{robust} recourse $\hat{x}^\text{cf}$ 
% (shown in \textcolor{cyan}{cyan}) 
% for input $x$ by anticipating the future shifted model $f(.;\theta')$.
% \\

% Figure \ref{fig:datashiftissue} illustrates this challenge of generating robust recourses. Let $f(.;\theta)$ denote an ML model's decision boundary. Given an input data point $x$, CF explanation methods generate
% a new counterfactual $\cf$ 
%(shown in \textcolor{orange}{yellow}) 
% as a recourse, which lies on the opposite side of decision boundary $f(.;\theta)$. However, as \textbf{new data} is made available, the ML model's decision boundary may 
% be retrained and get shifted (denoted as $f(.;\theta')$). 
% This shifted decision boundary invalidates the chosen recourse $\cf$ (as $x$ and $\cf$ lie on the same side of the shifted model $f(.;\theta')$). This is undesirable to an end-user because following recourse recommendations (as per $\cf$) would not lead to a favorable decision by the shifted ML model.
% \begin{wrapfigure}{r}{0.4\textwidth}
%   \begin{center}
%     \includegraphics[width=0.40\textwidth]{figs/robustcfnew.png}
%   \end{center}
%   \caption{\label{fig:datashiftissue}Illustration of the need for Robust Recourse Generation - shifts in data distribution leading to model shifts.}
%   \vspace{-10pt}
% \end{wrapfigure}
% Robust CF explanation methods address this problem by generating a \emph{robust} recourse $\hat{x}^\text{cf}$ 
%(shown in \textcolor{cyan}{cyan}) 
% for input $x$ by anticipating the future shifted model $f(.;\theta')$.

\noindent \textbf{Limitations of Prior Work.} To our knowledge, only two studies
\citep{upadhyay2021towards,nguyen2022robust} propose methods to generate robust recourses. Unfortunately, both these studies suffer from two major limitations. First, both methods are based on strong modeling assumptions which degrade their effectiveness at finding robust recourses (as we show in Section \ref{sec:experiment}). For example, \citet{upadhyay2021towards} assume that the ML model's decision boundary can be locally approximated via a linear function, and adopt LIME \citep{ribeiro2016lime} to find this linear approximation. However, recent works show that the local approximation generated from LIME is unfaithful \citep{laugel2018defining,rudin2019stop} and inconsistent \citep{slack2020fooling, alvarez2018robustness}. Similarly, \citet{nguyen2022robust} assumes that the underlying data distribution can be approximated via kernel density estimation \citep{bickel2009springer}. However, kernel density estimation suffers from the \emph{curse of dimensionality} \citep{bellman1959mathematical}, which performs exponentially worse with increasing dimensionality of data \citep{crabbe2013handling, nagler2016evading}. This limits its usability for estimating data distributions in real-world high-dimensional datasets.
% These simplifying assumptions result in several practical challenges (e.g., the use of LIME leads to inconsistency \citep{slack2020fooling} and unfaithfulness \citep{rudin2019stop}), which degrades the performance of these methods on real-world datasets. 

Second, these two techniques are post-hoc methods designed for use with proprietary black-box ML models whose training data and model weights are not available. However, %this black-box assumption is overly limiting in many real-world scenarios.
% This post-hoc paradigm has certain advantages, e.g., post-hoc explanation techniques are agnostic to the particulars of the ML model, and hence, they are generalizable enough to interpret any \emph{third-party} proprietary ML model. 
% However, we argue that in many real-world scenarios, the model-agnostic approach provided by post-hoc CF explanation methods is not desirable. 
with the advent of data regulations that enshrine the ``\emph{Right to Explanation}'' (e.g., EU-GDPR \citep{wachter2017counterfactual}), service providers are required by law to communicate both the decision outcome
(i.e., the ML model's prediction) and its actionable implications (i.e., a recourse for this prediction) to an end-user. 
In these scenarios, the post-hoc assumption is overly limiting, as service providers can build recourse models that leverage the knowledge of their ML model to generate higher-quality recourses. 
In fact, prior work \citep{guo2021counternet} has shown that post-hoc CF explanation approaches are unable to balance the cost-invalidity trade-off \citep{rawal2020can}, which is an important consideration in generating recourses. To date, very little prior work departs from the post-hoc paradigm; \cite{guo2021counternet} propose one such approach, unfortunately, it does not consider the robustness of generated recourses. \\

\noindent \textbf{Contributions.} We propose \textbf{\textit{Ro}}bust Re\textbf{\textit{Course}} Neural \textbf{\textit{Net}}work (or RoCourseNet), a novel framework for generating recourses which: (i) departs from the paradigm of post-hoc explainability in generating recourses; while (ii) optimizing the robustness of recourse explanations. %To that end, we propose an end-to-end training framework for simultaneously generating predictions and corresponding recourses (or CF explanations) that are robust to model shifts induced by distributional shifts in the training dataset. 
RoCourseNet presents four key contributions:

\begin{itemize}[leftmargin=*]
    \item (Formulation-wise) We formulate the robust recourse generation problem as a tri-level (min-max-min) optimization problem, which consists of two sub-problems: (i) a bi-level (max-min) problem which simulates a worst-case attacker to find an adversarially shifted ML model by explicitly simulating the \emph{worst-case data shift} in the training dataset; and (ii) an outer minimization problem which simulates an ML model designer who wants to generate robust recourses against this worst-case bi-level attacker. Unlike prior approaches, our bi-level attacker formulation explicitly connects shifts in the underlying data distribution to corresponding shifts in the ML model parameters. %TODO: Add a line why this better without overpromising too much, e.g., things that we can't justify in experiments
    \item (Methodology-wise) We propose \emph{RoCourseNet} for solving our tri-level optimization problem for generating robust recourses. RoCourseNet relies on two key ideas: (i) we propose a novel \emph{Virtual Data Shift (VDS)} algorithm to optimize for the inner bi-level (max-min) attacker problem, which results in an adversarially shifted model; and (ii) inspired by \citet{guo2021counternet}, RoCourseNet leverages a block-wise coordinate descent training procedure to optimize the robustness of generated recourses against these adversarially shifted models. Unlike prior methods \citep{upadhyay2021towards, nguyen2022robust}, our method requires no intermediate steps in approximating the underlying model or data distribution.
    %optimizes the inner bi-level (max-min) attacker problem by 
    %Inspired by \citet{guo2021counternet}, RoCourseNet leverages a (neural network) model based approach to jointly train the ML model and corresponding CF explanations, which enables RoCourseNet to better balance the cost-invalidity trade-off. In this way, we also depart from the prevalent post-hoc paradigm of generating CF explanations.
    %\item As the main contribution of this paper, 
    %In particular, we optimize for the outer minimization problem, we adopt a block-wise coordinate descent procedure \citep{guo2021counternet}. Further, to optimize for the inner bi-level max-min problem, we devise a novel \emph{Virtual Data Shift (VDS)} algorithm which finds an adversarially shifted model by explicitly simulating the \emph{worst-case data shift} in the training set. 
    \item (Experiment-wise) We conduct rigorous experiments on three real-world datasets to evaluate the robustness of several popular recourse generation methods under data shifts. Our results show that RoCourseNet generates highly robust CF explanations against data shifts, as it consistently achieves >96\% robust validity, outperforming state-of-the-art baselines by $\sim$10\%.%TODO: add a line on trade-off here.
    \item (Framework-wise) Finally, we extend the RoCourseNet training as a generalized robust training framework to be used with \emph{any} parametric post-hoc explanation method.
    % to improve the robustness of recourse explanation. 
    By applying the RoCourseNet training, we witness $\sim$25\% robust validity improvements to existing CF parametric methods.
    % provides us with a generalizable framework that can be applied with any parametric post-hoc CF explanation method to improve the robustness of recourses

    % \item We leverage adversarial training to optimize a complicated tri-level (min-max-min) optimization problem. 
    % \item RoCourseNet is built on top of CounterNet \citep{guo2021counternet}, an end-to-end framework (unlike post-hoc methods) that jointly trains predictions and CF explanations. Unlike post-hoc methods, CounterNet's prediction and CF explanations are jointly trained, which eliminates the misalignment issue prevalent in post-hoc methods. \jc{this sounds like contributions of CounterNet? What is unique about this new algorithm?}

    % \item We devise a novel \emph{metashift} algorithm to generate adversary shifted models. Unlike prior methods, \emph{metashift} generates shifted models by simulating the \emph{worst} data shifts in the training set. \jc{and why that is better than before?}
    % \item We propose rigorous experiments to evaluate the robustness of CF explanation methods under the distribution shift. \jc{and the results suggest what?}
\end{itemize}

\section{Related Work
}\label{sec:relatedwork}
\textbf{Counterfactual Explanation Techniques.}
A significant body of literature exists on CF explanation techniques, which focuses on generating recourses that lead to different (and often more preferable) predicted outcomes \citep{wachter2017counterfactual, verma2020counterfactual, karimi2020survey}. 
We categorize prior work on CF explanation techniques into \emph{non-parametric methods} \citep{wachter2017counterfactual, ustun2019actionable, mothilal2020explaining, van2019interpretable, karimi2021algorithmic, upadhyay2021towards,verma2020counterfactual,karimi2020survey}, which aim to find recourses without involving parameterized models, and \emph{parametric methods} \citep{pawelczyk2020learning, yang2021model, mahajan2019preserving, guo2021counternet}, which adopt parametric models (e.g., a neural network model) to generate recourses. 
In particular, our work is most closely related to CounterNet \citep{guo2021counternet}, which unlike post-hoc methods, jointly trains the predictive model and a CF explanation generator. This joint-training procedure leads to significantly better alignment between the generated predictions and corresponding CF explanations. 
\emph{However, all aforementioned CF explanation techniques (including CounterNet) do not optimize for robustness against adversarial model shifts. In contrast, we devise a novel tri-level adversarial training approach to ensure the robustness of CF explanations generated by RoCourseNet.}.\\
% However, CounterNet does not consider the commonly faced model shift problem, leading to CF explanations generated by CounterNet prone to getting invalidated when distributional data shifts occur (as shown in Section \ref{sec:experiment}). In contrast, we devise a novel tri-level adversarial training approach to ensure the robustness of CF explanations generated by RoCourseNet.

% which commonly occurs in practice due to shifts in the data distribution. As a result, the CF explanations generated by CounterNet are prone to getting invalidated when new data is made available to retrain (update) its predictive model. Contrastingly, we devise a novel three-level adversarial training approach to ensure that the CF explanation generated by RoCourseNet is robust to model shifts.
% our work builds on top of recent advances in adversarial training to ensure that the CF explanation generated by CounterNet is robust to model shifts.

\noindent\textbf{Robustness in Recourse Explanations.} Our method is closely related to the model shift problem in algorithmic recourse \citep{rawal2020can}, i.e., how to ensure that the generated recourse is robust to shifts in the underlying predictive model. However, existing approaches \citep{upadhyay2021towards,nguyen2022robust} rely on simplifying assumptions: (i) \citet{upadhyay2021towards} propose ROAR which relies on a locally linear approximation (via LIME \citep{ribeiro2016lime}) to construct a shifted model, which is known to suffer from inconsistency \citep{slack2020fooling,alvarez2018robustness} and unfaithfulness issues \citep{laugel2018defining,rudin2019stop}. 
(ii) Similarly, \citet{nguyen2022robust} propose RBR which assumes that kernel density estimators can approximate the underlying data distribution. In particular, RBR uses Gaussian kernels for multivariate density estimation, which suffers from the curse of dimensionality \citep{bellman1959mathematical, crabbe2013handling, nagler2016evading}. {In contrast, our work relaxes these assumptions by constructing adversarial shifted models via simulating the worst-case data shift, and conducting adversarial training for robust CF generation.}

Orthogonal to our work, \citet{pawelczyk2020counterfactual} analyze the \emph{model multiplicity} problem, which studies the validity of recourses under different ML models trained on the \emph{same} data, and \citet{black2022consistent} propose methods to ensure consistency under the model multiplicity setting. 
In addition, some prior work focus on ensuring robustness to small perturbations in the recourse features \citep{mishra2021survey, fokkema2022attribution, dominguez2022adversarial}. \\

\noindent\textbf{Adversarial training.}  
We leverage adversarial robustness techniques to protect ML models from adversarial examples~\citep{goodfellow2014explaining, madry2017towards, shafahi2019free, wong2019fast, chen2022efficient}.
%Some further improvements to PGD (e.g., free, fast, etc)... 
In addition, recent works \citep{geiping2021doesn, gao2022effectiveness} also leverage adversarial training to defend against data poisoning~\citep{huang2020metapoison} and backdoor attacks~\citep{saha2020hidden}.
In general, adversarial training solves a bi-level (min-max) optimization problem.
%Usually, the problem has a min-max structure, where the inner maximization problem aims at finding an adversary, and the outer minimization problem optimizes for this adversarial cases. %, e.g. mitigating the bias
In our work, we formulate RoCourseNet's objective as a tri-level (min-max-min) optimization problem, which we can decompose into a game played between a model designer and a worst-case (hypothetical) adversary.
%Similar to the adversarial training procedure, we solve this problem by first tackling the inner bi-level (max-min) problem, and optimizing for the adversarial model.
The inner worst-case data and model shifts are assumed to be generated by a bi-level worst-case attacker. The defender trains a robust CF generator against this bi-level attacker by following an adversarial training procedure.

\section{RoCourseNet: End-to-End Robust Recourse Generation} 
% \fj{For me, Section 3 + 4 are too much.  1) do not duplicate the contents 2) be concise 3) shortcut the story line}

RoCourseNet is an end-to-end training framework for simultaneously generating accurate predictions and corresponding recourses (or CF explanations) that are robust to model shifts induced by shifts in the training dataset. 
% At a high level, RoCourseNet jointly generates predictions and robust recourses by solving a tri-level optimization problem, which consists of two sub-problems: (i) a bi-level (max-min) problem which simulates a worst-case attacker by finding an adversarially shifted model; % by explicitly simulating the worst-case data shift in the training set;
% and (ii) an outer minimization problem that simulates a defender who wants to generate accurate predictions and robust recourses against this worst-case bi-level attacker. 
We describe the RoCourseNet framework in two stages. First, we discuss the attacker's problem: (i) we propose a novel bi-level attacker problem to find the worst-case data shift that leads to an adversarially shifted ML model; and (ii) we propose a novel Virtual Data Shift (VDS) algorithm for solving this bi-level attacker problem.
Second, we discuss the defender's problem: (i) we derive a novel tri-level learning problem based on the attacker's bi-level problem; and (ii) we propose the RoCourseNet training framework for optimizing this tri-level optimization problem, which leads to the simultaneous generation of accurate predictions and robust recourses.

%RoCourseNet is an end-to-end recourse generation framework that jointly optimizes the generation of accurate predictions and corresponding robust CF examples as part of a single end-to-end pipeline, which is built on top of CounterNet's architecture \citep{guo2021counternet}, and also adopts a block-wise coordinate descent procedure for training.
 
%However, unlike CounterNet (which does not optimize for robust recourses), RoCourseNet leverages adversarial training to ensure that the CF generator network (Figure \ref{fig:architecture}) produces \emph{robust} CF examples that maintain their validity under model shifts (induced by data shifts), such that $\cf$ is a robust CF example for input instance $x$ iff $f(\cf; \theta'_f) = f(\cf; \theta_f) = 1 - f(x; \theta_f)$, where $\theta_f$ and $\theta'_f$ represent the  parameters of original and shifted models, respectively.

%At a high level, RoCourseNet's adversarial training algorithm optimizes a robust CF generator by first constructing a worst-case model shift $f' \in \mathcal{F}$, and then optimizes the CF generator such that it generates CF examples $\cf$ which remain valid on the shifted model $f' \in \mathcal{F}$. Unlike prior work \cite{upadhyay2021towards}, the worst-case model shift is derived from an explicit data shift which perturbs the input instances (e.g., $x_{shift} = x + \delta$). This two-step procedure ensures that the trained CF generator produces CF examples which are highly robust to data shifts. Next, we elaborate on our algorithmic design. 

\subsection{Virtual Data Shift: Constructing Worst-case Data Shifts}
\label{sec:vds}
% ---------------- Clear version of model shift
% \textbf{Model shift as a bilevel optimization problem.} 
% To train a robust CF generator, we optimize against a virtual adversary who aims at creating a worst-case shifted model (denoted by its parameters $\theta'_f$) which can invalidate CF examples (i.e., $f(\cf; \theta'_{f}) \neq 1 - f(x; \theta_f)$).
% Crucially, we construct this worst-case shifted model by perturbing the training data.
% Constructing the shifted model from data shift is natural because, In reality, the model shift occurs due to the distributional shift in the training set. 
% As such, it is natural to simulate this model shift by considering the impact of data shift.
% Formally, this model shift problem can be considered as this bi-level optimization problem:
% \begin{equation}
%     \begin{split}
%         \label{eq:bi-level}
%         \operatorname{max}_{\delta \in \mathbb{B}(0, \epsilon)} &\ \mathbb{E}_{(x, y) \sim \mathcal{D}}\left[\mathcal{L}(f(\cf; \theta'_f^{*}(\delta)), 1 - f(x; \theta_f))\right] \\
%         s.t., &\ {\theta'_f}^{*}(\delta) = \operatorname{argmin}_{\theta'_f}
%         \mathbb{E}_{(x,y)\sim \mathcal{D}}\left[\mathcal{L}(f(x + \delta; \theta'_f), y)\right]
%     \end{split}
% \end{equation}
% where the outer problem maximizes the validity loss to construct a worst-case shifted data (i.e., $x_\text{shifted} = x + \delta$), and the inner problem minimizes the predictive loss on the shifted dataset to construct a worst-case shifted model $\theta'_f$.

% We first define some notation. 
We define a predictive model $f: \mathcal{X} \to \mathcal{Y}$. Let $\mathcal{D} = \{(x_i,y_i) \ | i \in \{1,\ldots,N\}\}$ represent our training dataset containing $N$ points. We denote $f(x,\theta)$ as the prediction generated by predictive model $f$ on point $x$, parameterized by $\theta$. Next, we denote $\theta$ and $\theta'$ as parameters of an (original) ML model $f(.;\theta)$ and its shifted counterpart $f(.;\theta')$, respectively. 
%$(x,y)\sim \mathcal{D}$ correspond to an input data point $x$ and ground truth label $y$ drawn i.i.d. from an underlying data distribution $\mathcal{D}$. 
% We denote $f(x,\theta)$ the prediction generated by ML model $f(.;\theta)$ on point $x$. 
Also, let $\cf$ denote a CF explanation (or recourse) for input point $x$. Finally, a recourse $\cf$ is \emph{valid} iff it gets an opposite prediction from the original data point $x$, i.e., $f(\cf;\theta)=1-f(x,\theta)$. On the other hand, a recourse $\cf$ is \emph{robustly valid} w.r.t. a shifted model $f(.;\theta')$ iff $\cf$ gets an opposite prediction from the shifted model (as compared to the prediction received by $x$ on the original model), i.e., $f(\cf;\theta')=1-f(x;\theta)$. This definition aligns with the notion of robustness to model shifts in the literature \citep{upadhyay2021towards}. Finally, $\mathcal{L}(.,.)$ represents a loss function formulation, e.g., binary cross-entropy, mean squared error, etc. \\

\noindent \textbf{Model Shift as an optimization problem.} To motivate the need of introducing our bi-level attacker problem, we first discuss an optimization problem for a worst-case attacker which directly perturbs model parameters to find an adversarially shifted model (denoted by $f(.;\theta'_{adv})$). The goal of the attacker is to find an adversarially shifted model which minimizes the robustness of the generated recourses. More formally, given our training dataset $\mathcal{D}$ and a CF explanation method (that can generate recourses $\cf$ for each $(x,y)\in \mathcal{D}$), the attacker's problem aims to find the worst-case shifted model $f(.;\theta'_{adv})$ which minimizes the robust validity of the generated recourses, i.e., $f(\cf; \theta'_{adv}) \neq 1 - f(x; \theta)$.
We can find an adversarial shifted model by solving Equation \ref{eq:1}.
\begin{equation}
    \label{eq:1}
    \theta'_{adv} = \operatorname{argmax}_{\theta' \in \mathcal{F}}\ 
    %\mathbb{E}_{(x,y)\sim \mathcal{D}}
    \frac{1}{N}\sum_{(x_i,y_i) \in \mathcal{D}}%\nolimits
    \bigg[
    \mathcal{L}\Big(f\left(\cf_i; \theta' \right), 1 - f\left(x_i; \theta\right)\Big)
    \bigg]
\end{equation}
where $\cf$ correspond to CF explanations of input $x$ produced by a CF generator, and $\mathcal{F} = \{\theta' \ | \ \theta + \delta_f \}$ denotes a plausible set of the parameters of all possible shifted models.

Unfortunately, it is non-trivial to construct a plausible model set $\mathcal{F}$ by directly perturbing the ML model's parameters $\theta$, especially when $f(\cdot; \theta)$ is represented using a neural network. Unlike a linear model, quantifying the importance of neurons is challenging \citep{leino2018influence, dhamdhere2018important}, which leads to difficulty in applying weight perturbations.
To overcome these challenges, prior work \citep{upadhyay2021towards} adopts a simplified linear model to approximate the target model, and perturbs this linear model accordingly. 
Unfortunately, this simplified local linear model introduces approximation errors into the system, which leads to poor performance (as shown in Section~\ref{sec:experiment}).
Instead of directly perturbing the model's weights, we explicitly consider a worst-case data shift, which then leads to an adversarial model shift.\\

% ---------------- Clear version of model shift above
\noindent\textbf{Data Shift as a bilevel optimization problem.} 
% \fj{add a figure}
% background, source of the non-robustness
%We first identify that non-robust CF explanations $\cf$ occur because the corresponding predictive model gets updated in response to distributional shifts in the training set.
% The invalidation of a CF example $\cf$ w.r.t. a (shifted) model occurs when this predictive model is changed due to the the distribution shifts in the training set.
% The non-robustness of the CF explanation $\cf$ occurs when the prediction model changes due to the distribution shifts in the training set.
We identify distributional data shifts as the fundamental cause of non-robust recourses. Unlike prior work, we propose a bi-level optimization problem for the attacker which explicitly connects shifts in the underlying training data to corresponding shifts in the ML model parameters. Specifically, in response to a shift in the training data (from $\mathcal{D}$ to $\mathcal{D}_{\text{shifted}}$), defenders update (or shift) their predictive model by optimizing the prediction loss:
% We denote the predictor learned from the shifted data $\mathcal{D}_{\text{shifted}}$ as $\theta'_{f}$, which optimize the predictive loss
\begin{equation}
\label{eq:predloss}
    \theta'_{opt} = \operatorname{argmin}_{\theta'}
        %\mathbb{E}_{(x,y)\sim \mathcal{D}_{\text{shifted}}}
        \frac{1}{N}\sum_{(x_i,y_i)\in\mathcal{D}_{\text{shifted}}}%\nolimits
        \bigg[\mathcal{L}\Big(f(x_i; \theta'), y_i\Big)\bigg].
\end{equation}
% motivation
Crucially, this updated ML model $f(\cdot,\theta'_{opt})$ (caused by the shifted data $\mathcal{D}_{\text{shifted}}$)
is the key to the non-robustness of recourses (i.e., $f(\cf; \theta'_{opt}) \neq 1 - f(x; \theta)$).
% introduce the robust optimization
Therefore, to generate robust recourses, we optimize against an adversary who creates a worst-case shifted dataset $\mathcal{D}^*_{\text{shifted}}$, such that the correspondingly updated model (found by solving Eq. \ref{eq:predloss}) minimizes the robust validity of CF examples $\cf$.
%who creates the worst-case shifted data $\mathcal{D}_{\text{shifted}}^*$, with a consideration of the trained predictive model which invalidates CF examples $\cf$.
%This shifted data $\mathcal{D}_{\text{shifted}}^*$ can be constructed from a set of the plausible data shifts $\Delta$ as $\mathcal{D}_{\text{shifted}} = \{(x+\delta, y)\ |\ \delta \in \Delta, (x,y) \in \mathcal{D}\}$.
% Let $\Delta$ denotes the set of plausible data shifts, we have $\mathcal{D}_{\text{shifted}} = \{(x+\delta, y)\ |\ \delta \in \Delta, (x,y) \in \mathcal{D}\}$.
% how we model the shift
% Here we consider the covariate shift, and $\Delta = \{\delta \in \mathbb{R}^n | ||\delta||_{p} \leq \epsilon \}$ for simplicity. Other design choice are shown in Appendix \fj{ref}.
Then, this data shift problem becomes a bi-level problem:
\begin{equation}
    \begin{split}
        \label{eq:bi-level}
        \boldsymbol{\delta}^{*} & =\mathop{\operatorname{argmax}}\limits_{\boldsymbol{\delta}, \forall \delta_i \in \Delta} 
        \ \frac{1}{N}\sum_{(x_i,y_i)\in\mathcal{D}}%\nolimits
        \bigg[\mathcal{L}\Big(f(\cf_i; \theta'_{opt}(\boldsymbol{\delta})), 1 - f(x_i; \theta)\Big)\bigg] \\
        s.t., & \ {\theta'}_{opt}(\boldsymbol{\delta}) = \operatorname{argmin}_{\theta'}\frac{1}{N}\sum_{(x_i,y_i)\in\mathcal{D}}%\nolimits
        \bigg[\mathcal{L}\Big(f(x_i + \delta_i; \theta'), y_i\Big)\bigg].
    \end{split}
\end{equation}
% \jc{directly use $\max_{\|\boldsymbol\delta\|_p \leq \epsilon }$? after s.t., prime location is wrong}
% \jc{notation-wise, in my experience, we would use boldsymbol for all vectors such as $\boldsymbol x$, $\boldsymbol \theta$, not sure whether we would like to change, but here you have both $\delta$ and $\boldsymbol\delta$ which seems a bit inconsistent.}
% The outer problem in Eq. \ref{eq:bi-level} maximizes the validity loss to construct worst-case data shift
where $\delta_i \in \Delta$ denotes the data shift for a single data point $x_i \in \mathcal{D}$,
% set of all plausible data shifts to a single data point. 
and
$\boldsymbol{\delta} = \{\delta_i\ |\ \forall (x_i, y_i) \in \mathcal{D}\}$
%$\boldsymbol{\delta} = \left[\delta_1, \delta_2, ..., \delta_m\right]$ 
denotes the data shift across all data points in the entire dataset. We define $\Delta$ as the $l_\infty$-norm ball $\Delta = \{\delta \in \mathbb{R}^n  \ | \  \|\delta\|_{\infty} \leq \epsilon \}$.
% (see $l_2$-norm based definitions in Appendix). 
Intuitively, the outer problem minimizes the robust validity to construct the worst-case data shift $\boldsymbol{\delta}^{*}$, and the inner problem learns a shifted model $f(.;\theta'_{opt})$ on the shifted dataset $\mathcal{D}^{*}_{\text{shifted}}$.
Once we get the optimal $\boldsymbol{\delta}^{*} = \{\delta^*_1, \delta^*_2,\ldots,\delta^*_N\}$ by solving Equation \ref{eq:bi-level}, the worst-case shifted dataset 
$\mathcal{D}^{*}_{\text{shifted}} = \{(x_i+\delta^{*}_i, y_i)\ |\ \forall (x_i,y_i) \in \mathcal{D}\}$.\\

\noindent\textbf{Virtual Data Shift (VDS).}
Unfortunately, solving Eq.~\ref{eq:bi-level} is computationally intractable due to its nested structure. To approximate this bi-level problem, we devise \emph{Virtual Data Shift (VDS)} (Algorithm~\ref{alg:metashift}), a gradient-based algorithm with an unrolling optimization pipeline. 
At a high level, VDS iteratively approximates the inner problem by unrolling $K$-steps of gradient descent for each outer optimization step. Similar unrolling pipelines are adopted in many ML problems with a bi-level formulation~\citep{shaban2019truncated, gu2022min}, e.g., meta-learning~\citep{finn2017model}, hyperparameter search~\citep{maclaurin2015gradient}, and poisoning attacks~\citep{huang2020metapoison}.

%When computing the outer problem's gradient, we look ahead the inner problem for a few forward steps and then back-propagate to the initial unrolling step. \fj{not sure whether it is clear or not}
% When calculating the gradient of adversarial loss (outer problem) with respect to data shift $\delta$ (Line~\ref{alg:line:meta-grad}), we look ahead the inner problem for a few forward steps and then back-propagate to the initial unrolling step.
%is computed via chain rule which back-propagates to the shifted model in the initial unrolling step. \fj{Redundant or not }
%, and optimizing for the outer problem via the approximated gradient implicitly considering $K$-unrolled-step back-propagation
%meta-gradient updates through these $K$-unrolling updates.
%In general, each iteration in the unrolling pipeline first takes \textit{K} unrolling steps to approximate the inner problem, then computes the outer update through $K$-unrolled-step back-propagation via chain rules. 

Algorithm~\ref{alg:metashift} layouts the \emph{VDS} algorithm which outputs the worst-case data shift $\boldsymbol{\delta}^{*}$, and the corresponding shifted model $f(.;\theta'_{opt})$. 
%iteratively tackles this bi-level problem.
VDS makes two design choices. % to effectively generate an adversarial shifted model. 
First, it uniformly randomizes the data shift $\boldsymbol{\delta} \sim \mathcal{U}(-\epsilon, + \epsilon)$, where $\boldsymbol{\delta}=\{\delta_1, \ldots, \delta_N \}$ (Line~\ref{alg:line:init}), following  practices of \citet{wong2019fast}. 
Uniform randomization is critical to adversarial model performance as it increases the smoothness of the objective function, leading to improved convergence of gradient-based algorithms \citep{chen2022efficient}.
% At the beginning, we copy the weights from the original model $\theta$ to the auxiliary shifted model $\theta'$, and uniformly randomize the data shift $\delta\sim \mathcal{U}(-\epsilon, + \epsilon)$, following  practices of \citet{wong2019fast}. \citet{chen2022efficient} pointed out that random initialization increases the smoothness of the optimization objective functions, leading to improved convergence of gradient-based algorithms.  
Then, we iteratively solve this bi-level optimization problem via $T$ outer attack steps. 
At each step, we first update the predictor $f(.;\theta')$ using the shifted data $\mathbf{x} + \boldsymbol{\delta}$ via $K$ unrolling steps of gradient descent (Line \ref{alg:line:unrolling}). 
% , where $\mathbf{x} = \{x_1, \ldots, x_N\}$ (Line \ref{alg:line:unrolling}). 
%Next, we adopt 
Next, similar to the fast sign gradient method (FSGM) \citep{goodfellow2014explaining}, we maximize the adversarial loss and project $\boldsymbol{\delta}$ into the feasible region $\Delta$ (i.e., $l_{\infty}$ norm ball; Line~\ref{alg:line:meta-grad}-\ref{alg:line:proj}). 
Crucially, when computing the gradient of adversarial loss (outer problem) w.r.t. data shift $\boldsymbol{\delta}$ (Line~\ref{alg:line:meta-grad}), we look ahead a few steps in the inner problem before back-propagating to the initial unrolling step.
This approach stems from applying $K$ unrolling steps of gradient descent, as opposed to full-blown gradient descent utill convergence. Note that the gradient w.r.t. $\delta$ depends on $\theta(\delta)$, where $\theta(\delta)$ is a function derived from LINE \ref{alg:line:unrolling_start}-\ref{alg:line:unrolling_end}.
\begin{algorithm}[t]
\caption{Virtual Data Shift (VDS)\label{alg:metashift}}
%\label{alg:adv_pred}
\begin{algorithmic}[1]
\State \textbf{Hyperparameters:} learning rates $\eta$, step size $\alpha$, \# of attacker steps $T$, \# of unrolling steps $K$
\State \textbf{Input:} model weights $\theta$, perturbation constraints $\epsilon$, batch $B=(\mathbf{x}, y)$, CF examples $\mathbf{x}^\text{cf}$
\State \textbf{Initialize: } virtual shifted model weights $\theta' = \theta$, $\boldsymbol{\delta}\sim \mathcal{U}(-\epsilon, + \epsilon)$ \label{alg:line:init}
\For{$i = 1\rightarrow T$ steps}
\For{$k = 1 \rightarrow K$ unroll steps}\label{alg:line:unrolling_start}
    %\For{each minibach $B$ in $\mathcal{D}$}
        % \State $\theta_f' =\theta_f' - \eta \cdot \nabla_{\theta_f'} \mathcal{L}_{(x, y) \in B}(f_{\theta_f'}(x + \delta)), y) $
        \State $\theta' \leftarrow \theta' - \eta \cdot \nabla_{\theta'} \mathcal{L}\left(f(\mathbf{x} + \boldsymbol{\delta}); \theta'), y\right) $ \label{alg:line:unrolling}
\EndFor\label{alg:line:unrolling_end}
\State $\boldsymbol{\delta} \leftarrow \boldsymbol{\delta} + \alpha \cdot sign\left(\nabla_{\delta} \mathcal{L}\left(f(\mathbf{x}^\text{cf}; \theta'), 1 - f(\mathbf{x}; \theta)\right)\right)$ \label{alg:line:meta-grad} %\fj{$x^{cf}$ or $g()$?}
% \State Project $\delta$ onto the feasible region ($\Delta = \{\delta \in \mathbb{R}^n\ |\ ||\delta||_{\infty} \leq \epsilon \}$). \label{alg:line:proj}
\State Project $\boldsymbol{\delta}$ onto the $l_\infty$-norm ball. \label{alg:line:proj}

\EndFor\\
\Return $\theta', \boldsymbol{\delta}$
\end{algorithmic}
\end{algorithm}
% \end{minipage}
% \end{wrapfigure}
% \hfill
% \begin{minipage}{.49\linewidth}
% \begin{algorithm}[H]
% \caption{Tri-level Robust CF Training}\label{alg:adv_training}
%     \begin{algorithmic}[1]
%     \State \textbf{Hyperparameters:} learning rates $\eta$, \# of epochs $N$, maximum perturbation constraints $E$
%     \State \textbf{Input: } dataset $(x, y) \in \mathcal{D}$
%     \State \textbf{Initialize: } $\theta = \{\theta_h, \theta_f, \theta_g\}$.
%     \For{epoch = $1\rightarrow N$}
%     \State $\epsilon = E \cdot \text{epoch} / N$ \Comment{Linearly schedule $\epsilon$.}
%     \For{each minibach $B$ in $\mathcal{D}$}
%     %\For{each minibach $B$ in $\mathcal{D}$}
%         \State $\theta = \theta - \eta \cdot \nabla_{\theta} {L}_1 $
%         \State $\theta_f' = \text{VDS}(B, \cf, \theta_f, \epsilon)$
%         \State $\theta_g = \theta_g - \eta \cdot \nabla_{\theta_g}\left(\lambda_2 \cdot {L}_2 + \lambda_3 \cdot {L}_3 \right)$
% \EndFor
% \EndFor
% \end{algorithmic}
%   \end{algorithm}
% \end{minipage}

\subsection{Block-wise Coordinate Descent with Adversarial Training}
\label{sec:adv_training}
\noindent \textbf{Choice of CF Explanation Technique.} Note that our bi-level attacker formulation assumes that CF explanations $\cf_i$ for all data points $x_i$ are provided as input to the VDS algorithm. Thus, a key design choice inside the RoCourseNet framework is the selection of an appropriate CF explanation technique, which can be used to generate recourses for input data points in Algorithm \ref{alg:metashift}. 

As mentioned in Section \ref{sec:relatedwork}, most existing CF explanation techniques follow the post-hoc paradigm, which makes them unsuitable for use inside the RoCourseNet framework for two reasons: (ii) \emph{misaligned motivations}: post-hoc CF explanation methods are mainly designed for use with proprietary black-box ML models whose training data and model weights are not available; instead, the VDS algorithm relies on having access to the training data. Thus, the motivations and use cases of VDS and post-hoc methods are misaligned. (ii) the post-hoc paradigm is overly limiting in many real-world scenarios. With the advent of data regulations that enshrine the ``\emph{Right to Explanation}'' (e.g., EU-GDPR \citep{wachter2017counterfactual}), service providers are required by law to communicate both the decision outcome
(i.e., the ML model's prediction) and its actionable implications (i.e., a recourse for this prediction) to an end-user. 
In these scenarios, the post-hoc assumption is overly limiting. Service providers can build specialized CF explanation techniques that leverage the knowledge of their specific ML model to generate higher-quality recourses. 

Motivated by these reasons, we choose CounterNet \citep{guo2021counternet} as our CF explanation model of choice inside the RoCourseNet framework, as that is an end-to-end approach that departs from the post-hoc explanation paradigm by jointly training predictions and recourses. In fact, \cite{guo2021counternet} show that CounterNet can better balance the cost-invalidity trade-off \citep{rawal2020can} than state-of-the-art post-hoc approaches. \\

\noindent \textbf{RoCourseNet Objective Function.} 
% We now describe how our bi-level attacker problem is plugged into an outer minimization problem to formulate a tri-level problem, which represents RoCourseNet's objective function. 
We describe how we combine the bi-level attacker problem with an outer minimization, which represents RoCourseNet's tri-level objective function.
Inspired by \cite{guo2021counternet}, RoCourseNet has three objectives: (i) high \emph{predictive accuracy} - we expect RoCourseNet to output accurate predictions; (ii) high \emph{robust validity} - we expect that generated recourses in RoCourseNet are robustly valid on shifted models\footnote{Note that robust validity also ensures validity on the original predictive model (i.e., $f(x; \theta) = 1 - f(\cf; \theta)$), as the worst-case model shift case encompasses the unshifted model case. We also empirically validate this design choice in Section\ref{sec:experiment}.}; (iii) low \emph{proximity} - we desire minimal changes required to modify input instance $x$ to corresponding recourse $\cf$. Given these objectives, RoCourseNet solves the following min-max-min problem to optimize parameters for its predictor $f(.;\theta)$ and recourse generator $g(.;\theta_g)$. Note that similar to \cite{guo2021counternet}, both $f(.;\theta)$ and $g(.;\theta_g)$ are neural networks.
% \begin{equation}
%     \label{eq:object}
%     \operatorname*{argmin}_{\mathbf{\theta}} 
%     \mathbb{E}_{\left(x, y\right) \sim D}
%     \bigg[ 
%     \lambda_1 \cdot \! \underbrace{\left(y- \hat{y}_{x}\right)^2}_{\text{Prediction Loss}\ (\mathcal{L}_1)} + 
%     \;\lambda_2 \cdot \;\; \underbrace{\left(\hat{y}_{x}- \left(1 - \hat{y}_{x'}\right)\right)^2}_{\text{Validity Loss}\ (\mathcal{L}_2)} \,+ 
%     \;\lambda_3 \cdot \!\! \underbrace{\left(x- x'\right)^2}_{\text{Proximity Loss}\ (\mathcal{L}_3)}
%     \bigg]
% \end{equation}
\begin{equation}
    \begin{aligned}
    \label{eq:adv_training}
    % \operatorname*{argmin}_{\mathbf{\theta}=\{\theta_h, \theta_m, \theta_g\}}
    &\operatorname*{argmin}_{\theta, \theta_g}
    % \mathbb{E}_{\left(x, y\right) \sim D}
    \Bigg(
    \frac{1}{N}\sum_{(x_i,y_i) \in \mathcal{D}}%\nolimits
    \bigg[ 
    \lambda_1 \cdot \!\! \underbrace{\mathcal{L}\Big(f(x_i; \theta), y_i\Big)}_{\text{Prediction Loss}\ ({L}_1)}+ \;
    \lambda_3 \cdot \!\!\!\!\!\!\! \underbrace{\mathcal{L}\Big(x_i, \cf_i\Big)}_{\text{Proximity Loss} \ ({L}_3)} \!\!\!\!\!\! \bigg] \\
    &+  \max_{\boldsymbol{\delta}, \forall \delta_i \in \Delta}
    % \mathbb{E}_{\left(x_i, y\right) \sim D}
    \frac{1}{N}\sum_{\left(x_i, y_i\right) \in \mathcal{D}}%\nolimits
    \bigg[ 
    \lambda_2 \cdot \underbrace{\mathcal{L}\Big(f\left(\cf_i; \theta^{\prime}_{opt}(\boldsymbol{\delta}) \right), 1 - f\left(x_i; \theta\right)\Big)}_{\text{Robust Validity Loss} \ ({L}_2)}
    \bigg]\Bigg) \\
    %& s.t., \mathcal{F} = \{\theta'_f \ | \  \operatorname{argmin}_{\theta'_f} \mathbb{E}_{(x,y)\in D}\left[ \mathcal{L}(f(x + \delta; \theta'_f), y)\right], \ \forall \ \delta \in \mathbb{B}(0, \epsilon)\}
    s.t., \ &{\theta'_{opt}}(\boldsymbol{\delta}) = \operatorname*{argmin}_{\theta'}
        % \mathbb{E}_{(x,y)\sim \mathcal{D}}
        \frac{1}{N}\sum_{\left(x_i, y_i\right) \in \mathcal{D}}%\nolimits
        \bigg[\mathcal{L}\Big(f(x_i + \delta_i; \theta'), y_i\Big)\bigg],\\
    &\cf_i = g(x_i; \theta_g).
    \end{aligned}
\end{equation}

% Formally, our robust training can be viewed as a three-level optimization problem:
% \begin{equation}\label{eq:adv_training}
%     \arg \min_{\theta} \mathcal{L}(f_{\theta_f}(X), Y) + \left[\max_{D_{\text{shift}}} \mathcal{L}_{\text{adv}} (X, Y; \theta_g,  \theta_f(X_{\text{shifted}}))\right]
% \end{equation}
% where $\theta'_f(X_{\text{shifted}}) = \arg\underset{\theta_f}{\min}\  \mathcal{L}(f_{\theta_f}(X_{\text{shifted}})), Y)$.

\begin{algorithm}[h]
\caption{Tri-level Robust CF Training}\label{alg:adv_training}
    \begin{algorithmic}[1]
    \State \textbf{Hyperparameters:} learning rates $\eta$, \# of epochs $N$, maximum perturbation $E$
    \State \textbf{Input: } dataset $(x, y) \in \mathcal{D}$
    \State \textbf{Initialize: } $\theta$.
    \For{epoch = $1\rightarrow N$}
    \State $\epsilon = E \cdot \text{epoch} / N$ \Comment{Linearly schedule $\epsilon$.}
    \For{each minibach $B$ in $\mathcal{D}$}
    %\For{each minibach $B$ in $\mathcal{D}$}
        \State $\theta \leftarrow \theta - \eta \cdot \nabla_{\theta} {L}_1 $ \label{alg:line:pred_loss}
        \State $\theta',\delta \leftarrow \text{VDS}(B, \cf, \theta, \epsilon)$ \label{alg:line:vds}
        \State $\theta_g \leftarrow \theta_g - \eta \cdot \nabla_{\theta_g}\left(\lambda_2 \cdot {L}_2 + \lambda_3 \cdot {L}_3 \right)$ \label{alg:line:val_loss}
\EndFor
\EndFor
\end{algorithmic}
\end{algorithm}

\noindent \textbf{RoCourseNet Training.} A common practice in solving a min-max formulation is to first solve the inner maximization problem, and then solve the outer minimization problem \citep{madry2017towards}. Thus, we solve Eq.~\ref{eq:adv_training} as follows (see Algorithm~\ref{alg:adv_training}): (i) To solve the inner bi-level problem, we find the worst-case model shift by solving Eq.~\ref{eq:bi-level} using \emph{VDS} (Algorithm~\ref{alg:metashift}).
(ii) To solve the outer minimization problem, we adopt block-wise coordinate descent optimization by distributing gradient descent backpropagation on the objective function into two stages (as suggested in \citep{guo2021counternet}) -- at stage one, we optimize for predictive accuracy, i.e., ${L}_1$ in Eq.~\ref{eq:adv_training} (Line~\ref{alg:line:pred_loss}), and at stage two, optimize the quality of CF explanations, i.e., $\lambda_2 \cdot {L}_2 + \lambda_3 \cdot {L}_3$ in Eq.~\ref{eq:adv_training} (Line~\ref{alg:line:val_loss}). Note that in Line~\ref{alg:line:val_loss}, the gradient of $\lambda_2 \cdot {L}_2 + \lambda_3 \cdot {L}_3$ is calculated using the adversarially shifted model parameters $\theta'$ found by VDS in Line \ref{alg:line:vds}. \cite{guo2021counternet} show that this block-wise coordinate descent algorithm efficiently optimizes the outer minimization problem by alleviating the problem of divergent gradients.

%addresses the issue of poor convergence of solving the outer minimization problem as-is via gradient descent.

%optimizes the outer minimization problem by alleviating the divergent gradient issue (which leads to poor convergence of solving the outer minimization problem as-is via gradient descent).

%, i.e., the gradient directions of prediction loss (${L}_1$) and validity loss (${L}_2$) moves in different directions, which leads to poor convergence of solving the outer minimization problem as-is via gradient descent (proven in \citet{guo2021counternet}).

% \citet{guo2021counternet} pointed out that directly optimizing the objective function as-is via gradient descent (i.e., solving the minimization problem in Eq.~\ref{eq:adv_training}) suffers from the divergent gradient issue (i.e., the gradient direction of prediction loss and validity loss moves in different directions) which leads to poor convergence of training. On the other hand, a block-wise coordinate descent algorithm alleviates this problem by distributing the objective function into two stages without suffering from gradients of different objectives canceling out with each other.

In addition, we linearly increase the perturbation constraints $\epsilon$ for improved convergence of our robust CF generator. Intuitively, linearly increasing $\epsilon$ values correspond to increasingly strong adversaries. Prior work in curriculum adversarial training \citep{cai2018curriculum, wang2019convergence} suggests that adaptively adjusting the strength of the adversary improves the convergence of adversarial training (we verify this in Section~\ref{sec:experiment}). %Empirically, we observe that adopting curriculum adversarial training boosts the performance of adversarial training (see Section~\ref{sec:experiment}).

\section{Experimental Evaluation}
\label{sec:experiment}

%We set values of $K=2$ (same as \citep{huang2020metapoison}) and learning rate $\alpha = 2.5 \times \epsilon / T$ (based on \citep{madry2017towards}), where the perturbation budget $\epsilon$ is chosen adaptively as described above. The unrolling learning rate $\eta$ is separately tuned while training RoCourseNet (see Appendix).

% \subsection{Experiment Setup}

%Since, to our knowledge, FASTAR is the first approach to generate amortized ARs for black-box models, there exist no previous approaches which we can directly compare against. Nevertheless, we compare FASTAR to several previous popular AR generating approaches.

% We primarily focus our evaluation on heterogeneous tabular datasets for binary classification problems (which is the most common and reasonable setting for CF explanations \citep{verma2020counterfactual, stepin2021survey}). However, CounterNet can be applied to multi-class classification settings, and it can also be adapted to work with other modalities of data, e.g., images, etc. (as shown in Appendix~\ref{sec:app-muliclass}, \ref{sec:app-image}).\\

\noindent\textbf{Baselines.} To our knowledge, RoCourseNet is the first method that optimizes an end-to-end model for generating predictions and robust recourses. Hence, there exist no previous approaches which we can directly compare against. Nevertheless, we compare RoCourseNet against four state-of-the-art baselines.
\begin{itemize}[leftmargin=*]
    \item \textsc{VanillaCF} \citep{wachter2017counterfactual} is a popular post-hoc non-parametric method optimizing validity and proximity.
    \item \textsc{CounterNet} \citep{guo2021counternet} is the first end-to-end model for simultaneously generating predictions and recourses, which inspires the development of RoCourseNet.
    \item \textsc{ROAR-LIME} \citep{upadhyay2021towards} generates robust recourses by perturbing parameters of locally approximated linear models.
    \item \textsc{RBR} \citep{nguyen2022robust} generates robust recourses from a Gaussian kernel.
\end{itemize}

% , there exist no previous approaches which we can directly compare against. Hence, we compare  Nevertheless, we compare RoCourseNet with two post-hoc robust recourse generation methods: (i) \textsc{ROAR-LIME} \citep{upadhyay2021towards}; and (ii) \textsc{RBR} \citep{nguyen2022robust}. In addition, to remain consistent with baselines established in prior work \citep{upadhyay2021towards,nguyen2022robust}, we also compare against \textsc{VanillaCF} \citep{wachter2017counterfactual}, a popular post-hoc method. Finally, we also compare against \textsc{CounterNet} \citep{guo2021counternet} since our approach is inspired by their end-to-end paradigm of generating recourses. 

Other than \emph{CounterNet}, all remaining baselines require a trained ML model as an input. Similar to \citep{guo2021counternet}, we train a neural network model as the base ML model for all baselines. For each dataset, we separately tune hyperparameters via grid search. \\
% (see Appendix~\ref{sec:app-implementation}).\\

\begin{table}[t]
\caption{\label{tab:summary_data}Summary of Datasets used for Evaluation. Each dataset consists of $k$ subsets. \emph{Size} represents the number of data points for the entire dataset.}
\begin{tabular}{l|lccc} 
\toprule
\textbf{Dataset} & \textbf{Size} & {$\boldsymbol{k}$} & \textbf{\#Continuous} & \textbf{\#Categorical} \\ \midrule\midrule
Loan & 449,152 & 16 & 7 & 5 \\
German Credit & 2,000 & 2 & 6 & 3 \\
Student & 649 & 2 & 2 & 13 \\
\bottomrule
\end{tabular}
\end{table}

\noindent\textbf{Datasets.}
To remain consistent with prior work on robust recourses \citep{upadhyay2021towards,nguyen2022robust}, we evaluate RoCourseNet on three benchmarked real-world datasets. Table~\ref{tab:summary_data} summarizes these three datasets.
\begin{itemize}[leftmargin=*]
    \item \emph{Loan}~\citep{li2018should} captures \emph{temporal shifts} in loan application records. It predicts whether a business defaulted on a loan (\verb|Y=1|) or not (\verb|Y=0|). This large-sized dataset (i.e., $\sim$450k data points) has loan approval records across the U.S. during 1994 to 2009 (i.e., $\mathcal{D}_{\text{all}} = \{\mathcal{D}_1, \mathcal{D}_2, ..., \mathcal{D}_k \}$, where each subset $\mathcal{D}_i$ corresponds to a particular year, and $k=16$ is the total number of years). %We adopt Random forests \citep{breiman2001random} to select the 12 most predictive features, including 7 continuous and 5 categorical features. 
    \item \emph{German Credit}~\citep{asuncion2007uci} captures \emph{data correction shifts}. It predicts whether the credit score of a customer is good (\verb|Y=1|) or bad (\verb|Y=0|). This dataset has 2,000 data points with two versions; each version contains 1,000 data points (i.e., $\mathcal{D}_{\text{all}} = \{\mathcal{D}_1, \mathcal{D}_2\}$, where $\mathcal{D}_1$, $\mathcal{D}_2$ corresponds to the original and corrected datasets, respectively). %It contains 6 continuous and 3 categorical features.
    \item Finally, we use the \emph{Student} dataset~\citep{cortez2008using}, which captures \emph{geospatial shifts}. It predicts whether a student will pass (\verb|Y=1|) or fail (\verb|Y=0|) the exam. It contains 649 student records in two places (i.e., $\mathcal{D}_{\text{all}} = \{\mathcal{D}_1, \mathcal{D}_2\}$, and $\mathcal{D}_1$, $\mathcal{D}_2$ represent a particular school).\\ %It contains 2 continuous and 14 categorical features.
%It recorded 649 students
\end{itemize}

\begin{figure}[t]
    \centering
    \includegraphics[width=0.8\columnwidth]{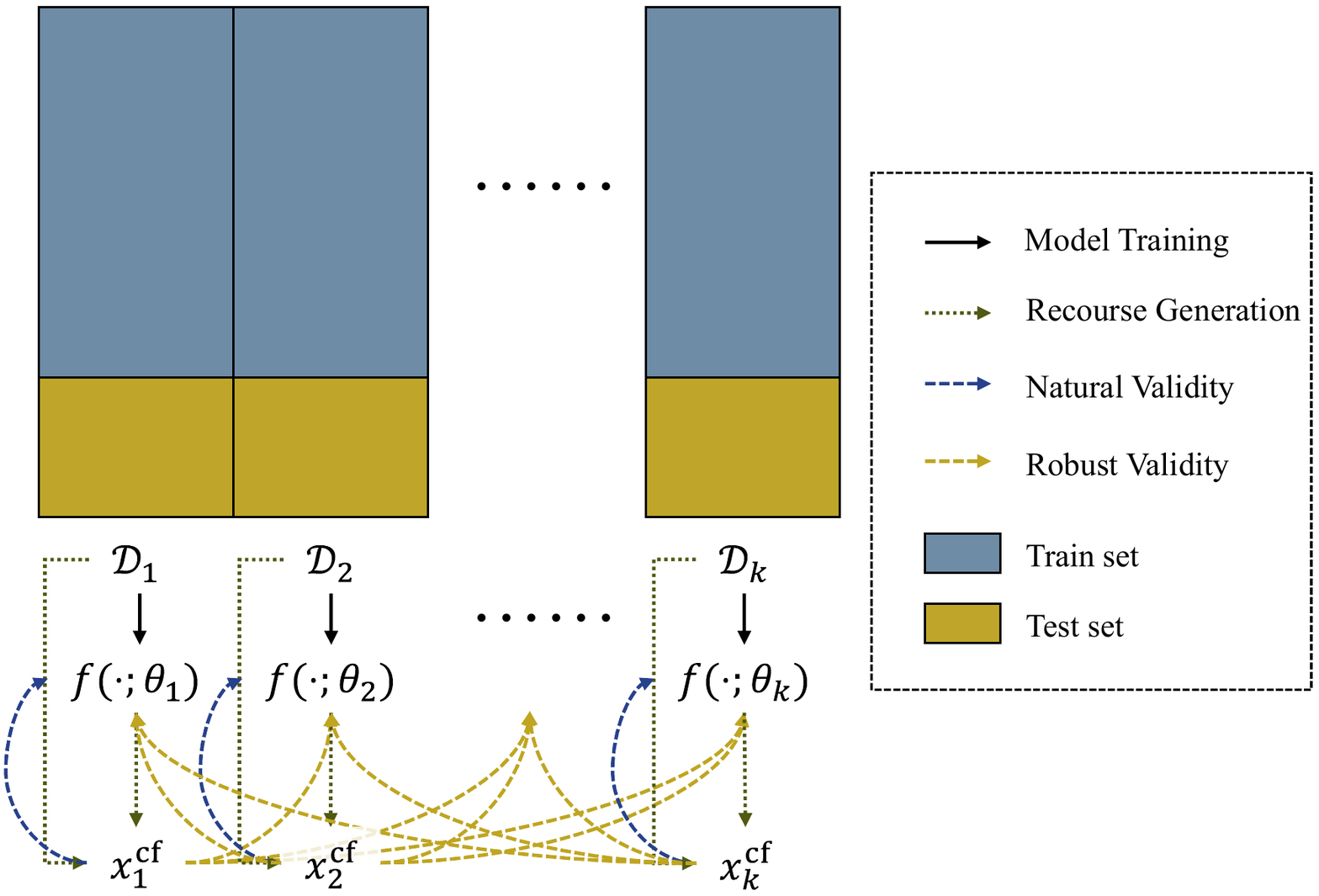}
    \caption{Illustration of evaluating the recourse robustness under the distributional shift.}
    \label{fig:procedure}
    \vspace{-5pt}
\end{figure}

\noindent\textbf{Evaluation Procedure \& Metrics.} 
Figure~\ref{fig:procedure} illustrates the evaluation procedure of the experiment.
Each dataset is partitioned into $k$ subsets $\mathcal{D}_{\text{all}} = \{\mathcal{D}_1, \mathcal{D}_2, ..., \mathcal{D}_k \}$. This partitioning enables us to create $k$ original datasets  $\mathcal{D}_i \in \mathcal{D}_{\text{all}}$, and $k$ corresponding shifted datasets, $\mathcal{D}_{\text{shifted},i} = \mathcal{D}_{\text{all}}\setminus \mathcal{D}_i$, 
% We denote the original dataset as $\mathcal{D}_i$, and the shifted datasets as $\mathcal{D}_{\text{shifted},i} = \mathcal{D}_{\text{all}}\setminus \mathcal{D}_i$, 
where each shifted dataset $\mathcal{D}_{\text{shifted},i}$ contains all subsets of the dataset $\mathcal{D}_{\text{all}}$ except for the original dataset $\mathcal{D}_i$.
Next, we do a train/test split on each $\mathcal{D}_i \in \mathcal{D}_{\text{all}}$ (i.e., $\mathcal{D}_i = \{\mathcal{D}^\text{train}_i, \mathcal{D}^\text{test}_i\}$). 
% Next, we split each subset $\mathcal{D}_i$ into a train set and a hold-out test set (i.e., $\mathcal{D}_i = \{\mathcal{D}^\text{train}_i, \mathcal{D}^\text{test}_i\}$). 
We train a separate model (i.e., the entire models for \emph{RoCourseNet} and \emph{CounterNet}, and predictive models for other baselines) on each train set $\mathcal{D}^\text{train}_i,~\forall i \in \{1,..., k\}$.
%For post-hoc methods (such as VanillaCF and Roar-LIME), a predictive model is trained for each sub-dataset $\mathcal{D}_i$. For CounterNet and ReCourseNet, we train each of these models on each sub-dataset. 
Then, we use the model trained on $\mathcal{D}_i^\text{train}$ to generate recourses on the hold-out sets $\mathcal{D}_i^\text{test},~\forall i \in \{1,..., k\}$.
Finally, we evaluate the robustness (against the model shift) of recourses generated on $\mathcal{D}_i^\text{test},~ \forall i \in \{1,\ldots, k\}$ as follows: for each recourse $\cf$ (corresponding to an input instance $x$ in $\mathcal{D}_i^\text{test}$), we evaluate its robust validity by measuring the fraction of shifted models (i.e., $k-1$ models trained on all shifted training sets) on which $\cf$ remains \emph{robustly valid} (see definitions in Section \ref{sec:vds}).
%  $\mathcal{D}_{i}^\text{train}\ \forall i \in \{1,k\} \setminus i$
% all predictive models trained on the shifted data (i.e., $\{\theta'_f \ | \ \theta'_f = \operatorname{argmin}_{\theta'_f} \mathbb{E}_{(x,y) \in \mathcal{D}_j^\text{train}} [\mathcal{L}(f(x; \theta'_f), y) ], \ \forall \ \mathcal{D}_j \in  \mathcal{D}_{\text{shifted},i}^\text{test}\}$).

Finally, we use three metrics to evaluate a CF explanation:
(i) \emph{Validity} is the fraction of valid CF examples on the original model $f(.;\theta)$.
(ii) \emph{Robust validity} is the fraction of robustly valid CF examples on a \emph{shifted} predictive model $f(.;\theta')$. We calculate the robust validity on all possible shifted models (as described above).
(iii) \emph{Proximity} is the $l_1$ distance between the input and the CF example.
We report the averaged results across all subsets $\mathcal{D}_{all}$ (see Table~\ref{tab:cf}).

\begin{table*}[th]
\setlength{\tabcolsep}{3.5pt}
\footnotesize
% \footnotesize
% \tiny
\centering
\caption{\label{tab:cf}\textbf{Evaluation of recourse robustness under model shift.} It is desirable for recourse methods to have \emph{low} \fcolorbox{cood}{cid}{proximity} ({prox.}) with \emph{high} \fcolorbox{cood}{cood}{validity} ({Val.}) and \emph{high} \fcolorbox{cood}{ccon}{robust validity} ({Rob-Val.}). }
\begin{tabular}{@{}l|rcc|rcc|rcc@{}}
\toprule
{\footnotesize\bf Methods}    & \multicolumn{3}{c|}{\footnotesize\bf Loan}   & \multicolumn{3}{c|}{\footnotesize\bf German Credit}  & \multicolumn{3}{c}{\footnotesize\bf Student}                                                             \\ %\cmidrule(lr){3-5}
& \cellcolor{cid} {\footnotesize Prox.} & \cellcolor{cood} {\footnotesize Val.} &  \cellcolor{ccon} {\footnotesize Rob-Val.} & \cellcolor{cid} {\footnotesize Prox.} & \cellcolor{cood} {\footnotesize Val.} &  \cellcolor{ccon} {\footnotesize Rob-Val.} & \cellcolor{cid} {\footnotesize Prox.} & \cellcolor{cood} {\footnotesize Val.} &  \cellcolor{ccon} {\footnotesize Rob-Val.} \\ \midrule\midrule
\textsc{\footnotesize VanillaCF}   & \cell{7.390 \\ $\pm$ 1.860}  & \cell{0.942 \\ $\pm$ 0.026}  &  \cell{0.885 \\ $\pm$ 0.121} &  \cell{\bf 4.635 \\ \bf $\pm$ 0.197} &    \cell{0.940 \\ $\pm$ 0.011}       &   \cell{0.772 \\  $\pm$  0.008}   &  \cell{15.236 \\ $\pm$ 0.383} &    \cell{0.915 \\ $\pm$ 0.056}       &   \cell{0.673 \\ $\pm$  0.006}              \\ \midrule
\textsc{\footnotesize CounterNet}  & \cell{6.746 \\ $\pm$ 0.723}  &   \cell{0.964 \\ $\pm$ 0.085} &    \cell{0.639 \\ $\pm$ 0.222} & \cell{5.719 \\ $\pm$ 0.130}  &  \cell{0.960 \\ $\pm$ 0.028}   &  \cell{0.706 \\ $\pm$ 0.074} & \cell{18.619 \\ $\pm$ 0.131} &  \cell{0.967 \\ $\pm$ 0.033}   &  \cell{0.859 \\ $\pm$ 0.071}      \\ \midrule
\textsc{\footnotesize Roar-Lime} & \cell{7.648 \\ $\pm$ 1.951} &   \cell{0.934 \\ $\pm$ 0.024}   &   \cell{0.918 \\ $\pm$ 0.066} &  \cell{4.862 \\ $\pm$ 0.117} &  \cell{0.910 \\ $\pm$ 0.0255}   & \cell{0.792 \\ $\pm$ 0.052} & \cell{\bf11.931 \\ \bf $\pm$ 1.396} &   \cell{0.967 \\ $\pm$ 0.026}   &  \cell{0.820 \\ $\pm$ 0.075}   \\ \midrule
\textsc{\footnotesize RBR} & \cell{11.71 \\ $\pm$1.633} &  \cell{0.824 \\ $\pm$ 0.130} & \cell{0.821 \\ $\pm$ 0.132} & \cell{6.005 \\ $\pm$ 2.099} &  \cell{0.524 \\ $\pm$ 0.148} &  \cell{0.586 \\ $\pm$ 0.046} & \cell{26.255 \\ $\pm$ 3.089} &  \cell{0.660 \\ $\pm$ 0.014} & \cell{0.611 \\ $\pm$ 0.073} \\ \midrule
\rowcolor[HTML]{EFEFEF}
\textsc{\footnotesize RoCourseNet} &    \cell{\bf6.611 \\\bf $\pm$ 0.418} &  \cell{\bf 0.996 \\\bf $\pm$ 0.002} &    \cell{\bf 0.969 \\\bf $\pm$ 0.106}  & \cell{6.903 \\ $\pm$ 0.250} &      \cell{\bf0.978 \\\bf $\pm$ 0.002} &    \cell{\bf0.964 \\\bf $\pm$ 0.008}   &   \cell{16.508 \\ $\pm$ 0.281} &   \cell{\bf1.000 \\\bf $\pm$ 0.000} &    \cell{\bf1.000 \\\bf $\pm$ 0.000}  \\

                                        \bottomrule

\end{tabular}
\end{table*}

\subsection{Experimental Results}

\noindent \textbf{Validity \& Robust Validity.} 
% \fj{Should we includes more details, like feature we used in Appendix}
Table~\ref{tab:cf} compares the validity and robust validity achieved by RoCourseNet and other baselines. RoCourseNet achieves the highest \emph{robust validity} on each dataset - it outperforms ROAR-LIME by 10\% (the next best performing baseline), and consistently achieves at least 96.5\% robust validity. This illustrates RoCourseNet's effectiveness at generating highly robust recourses. Also, RoCourseNet achieves the highest \emph{validity} on each dataset. This result shows that optimizing the worst-case shifted model (i.e., $L_2$ in Eq.~\ref{eq:adv_training}) is sufficient to achieve high validity, without the need to explicitly optimize for an additional validity loss.\\

%This table shows that RoCourseNet outperforms all baselines in terms of both validity and robust validity. In particular, RoCourseNet consistently generates CF examples with 91\% robust validity, which shows that the adversarially trained RoCourseNet generates highly robust CF examples when the underlying predictive model shift occurs. Moreover, RoCourseNet achieves $\sim$10\% higher robust validity than Roar-LIME (which optimizes for worst-case model shifts). This result accentuates the importance of considering the data shift, since failing to consider the impact data shift (such as ROAR-LIME) leads to degraded results.

%In addition, consistently generates CF examples with more than 98.5\% validity, which is $\sim$6\%, $\sim$5\%, $\sim$3\% higher than VanillaCF, ROAR-LIME, and CounterNet, respectively. This result shows that optimizing the worst-case shifted model (i.e., $L_2$ in Eq.~\ref{eq:adv_training}) is sufficient to achieve high validity, without the need to explicitly optimize for an additional validity loss on the original model.

% This table shows that RoCourseNet is the only method that consistently generates CF examples with 98\% natural validity and 91\% robust validity. Notably, ReCourseNet achieves $\sim$10\% higher robust validity than Roar-LIME (the closest competitor which optimizes for worst-case linear model shift). Moreover, RoCourseNet achieves $\sim$5\% higher natural validity than Roar-LIME. This is expected due to Roar-LIME's adoption of a linear approximation as the target model, which leads to poor validity performance.

\noindent\textbf{Proximity.} Table~\ref{tab:cf} compares the proximity achieved by RoCourseNet and baselines. In particular, RoCourseNet performs exceedingly well on the Loan application dataset (our largest dataset with $\sim$450k data points), as it is the second-best method in terms of proximity (just behind CounterNet), and outperforms all our post-hoc baseline methods (RBR, ROAR-LIME, and VanillaCF). Perhaps understandably, RoCourseNet achieves poorer proximity on the German Credit and Student dataset, given that the limited size of these datasets (less than 1000 data points) precludes efficient training.\\

% \vspace{-4pt}
\begin{figure*}[ht!]
    \centering
    \includegraphics[width=0.8\textwidth]{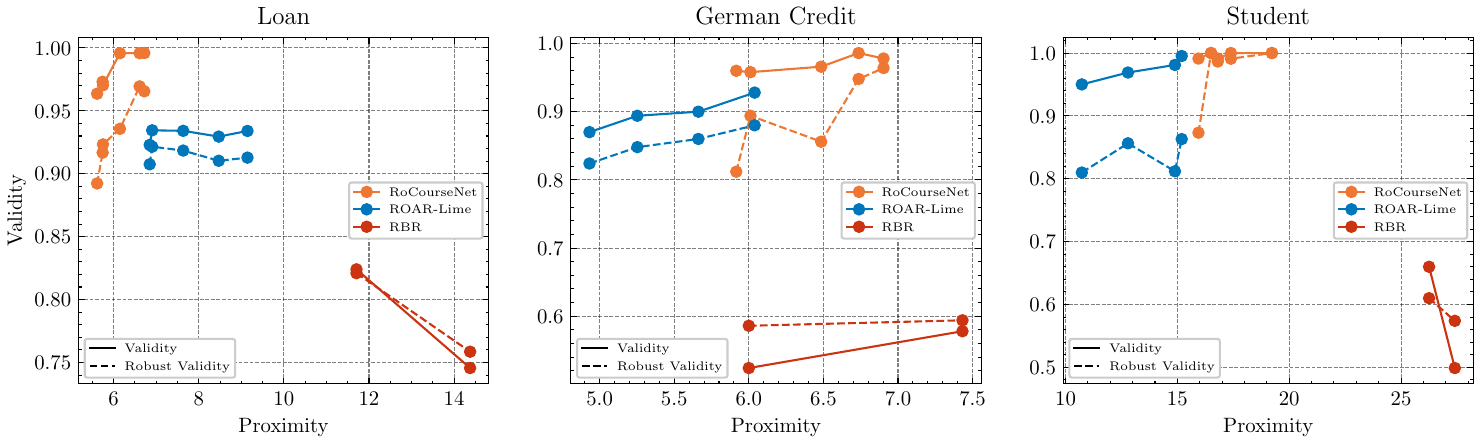}
    % \caption{Tradeoff figure. (Red line represents RoCourseNet, and blue line represents ROAR)}
    \caption{Pareto frontiers of the cost-invalidity trade-off for \textsc{Roar-Lime}, \textsc{RBR}, and \textsc{RoCourseNet}. Methods located in the upper-left region are preferable, as they exhibit a favorable balance between cost and invalidity. On the \emph{Loan} (left) dataset, \textsc{RoCourseNet} exhibits a superior balance between cost and invalidity compared to \textsc{Roar-Lime} and \textsc{RBR}. On the \emph{German Credit} (middle) and \emph{Student} (right) dataset, \textsc{RoCourseNet} achieves high normal and robust validity at the cost of proximity score.}
    \label{fig:pareto}
    \vspace{-5pt}
\end{figure*}

\noindent\textbf{Cost-Validity Trade-Off.} We compare RoCourseNet, ROAR-LIME, and RBR (three recourse methods explicitly optimizing for distributional shift) in their trade-off between the cost (measured by proximity) and their original and robust validity \citep{rawal2020can}. For each method, we plot the Pareto frontier of the cost-validity trade-off as follows: (i) For RBR, we obtain the Pareto frontier by varying the ambiguity sizes $\epsilon_1$, $\epsilon_2$ (i.e., hyperparameters of RBR \cite{nguyen2022robust}); (ii) For ROAR-LIME and CounterNet, we obtain the Pareto frontier by varying the trade-off hyperparameter $\lambda$ that controls the proximity loss term in their respective loss functions (e.g., $\lambda_3$ in Equation \ref{eq:adv_training}). Figure~\ref{fig:pareto} shows that on the large-sized Loan dataset, RoCourseNet's Pareto frontiers either dominate or are comparable to frontiers achieved by ROAR-LIME and RBR. On the other hand, we observe a clear trade-off on the German Credit and Student datasets, where RoCourseNet (and other methods) can increase their normal and robust validity, but only at the cost of a poorer proximity score. This is consistent with prior literature, which shows that proximity needs to be sacrificed to achieve higher validity \citep{nguyen2022robust}.

\begin{figure*}[ht]
     \centering
     \begin{subfigure}[h]{0.3\textwidth}
         \centering
         \includegraphics[width=0.9\textwidth]{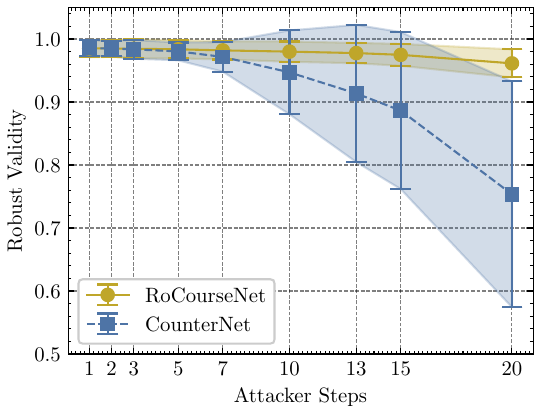}
         \caption{The number of attacker steps $T$ vs attacker effectiveness ($\downarrow$)}
         \label{fig:n_steps_attacker}
     \end{subfigure}
     \hfill
     \begin{subfigure}[h]{0.3\textwidth}
         \centering
         \includegraphics[width=0.9\textwidth]{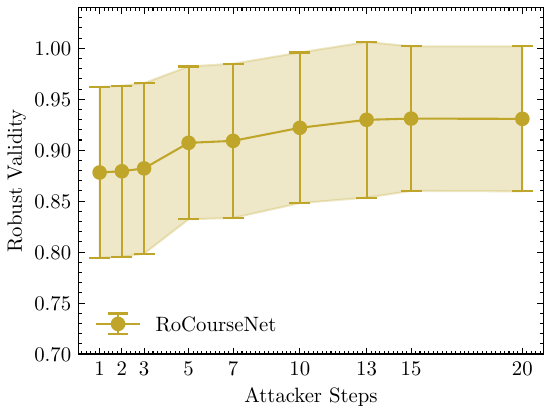}
         \caption{The number of attacker steps $T$ vs robust validity ($\uparrow$).}
         \label{fig:n_steps_rob_val}
     \end{subfigure}
     \hfill
     \begin{subfigure}[h]{0.3\textwidth}
         \centering
         \includegraphics[width=0.9\textwidth]{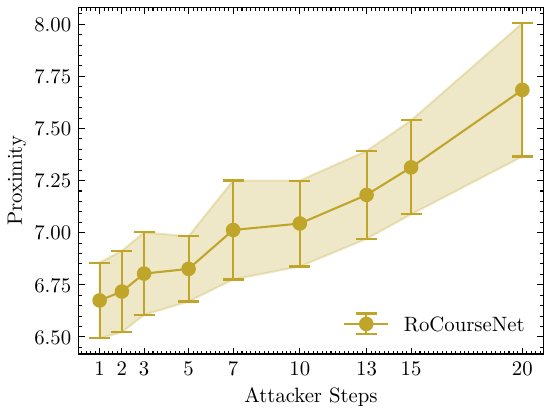}
         \caption{The number of attacker steps $T$ vs proximity ($\downarrow$).}
         \label{fig:n_steps_prox}
     \end{subfigure}
    \label{fig:n_steps_robustness}
     \begin{subfigure}[h]{0.3\textwidth}
         \centering
         \includegraphics[width=0.9\textwidth]{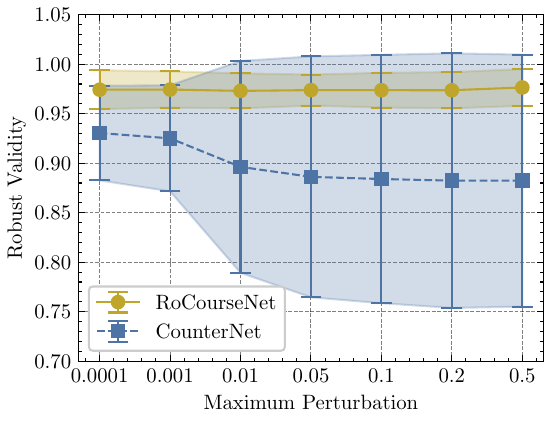}
         \caption{Maximum perturbation $E$ vs attacker effectiveness ($\downarrow$).}
         \label{fig:eps_attacker}
     \end{subfigure}
     \hfill
     \begin{subfigure}[h]{0.3\textwidth}
         \centering
         \includegraphics[width=0.9\textwidth]{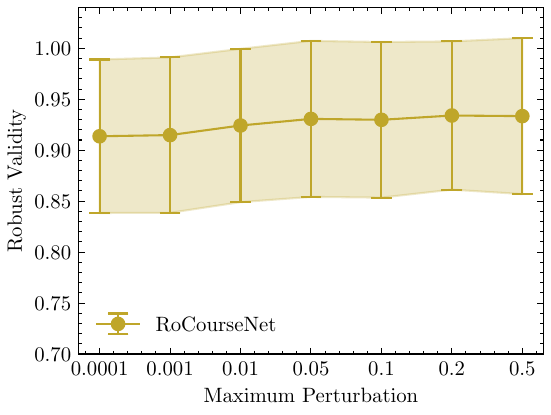}
         \caption{Maximum perturbation $E$ vs robust validity ($\uparrow$).}
         \label{fig:eps_rob_val}
     \end{subfigure}
     \hfill
     \begin{subfigure}[h]{0.3\textwidth}
         \centering
         \includegraphics[width=0.9\textwidth]{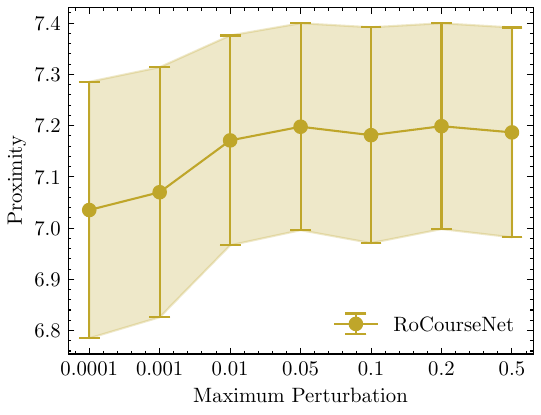}
         \caption{Maximum perturbation $E$ vs proximity ($\downarrow$).}
         \label{fig:eps_prox}
     \end{subfigure}
    \caption{The impact of the number of attacker steps $T$ (\ref{fig:n_steps_attacker}-\ref{fig:n_steps_prox}) and max-perturbation $E$ (\ref{fig:eps_attacker}-\ref{fig:eps_prox}) to robustness on the \emph{Loan} dataset ($\uparrow$ or $\downarrow$ means that higher or lower value is preferable, respectively).} %(i) \ref{fig:n_steps_attacker} and \ref{fig:eps_attacker} depict the role of $T$ (when $E=0.1$) and $E$ (when $T=15$) in the attacker settings, respectively. (ii) \ref{fig:n_steps_rob_val}, \ref{fig:n_steps_prox} and \ref{fig:eps_rob_val}, \ref{fig:eps_prox} describe the role of $T$ and $E$ in the defense setting, respectively.}
    \label{fig:robustness}
\end{figure*}

\begin{figure}[ht]
    \centering
    \includegraphics[width=0.95\columnwidth]{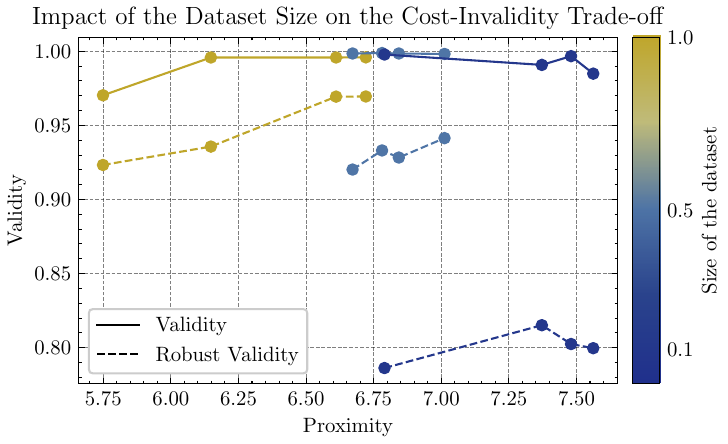}
    \caption{The influence of dataset size on the cost-invalidity trade-off. We analyze \textsc{RoCourseNet} on various fractions of the \emph{Loan} Dataset. The Pareto Frontier plot reveals improvement in both normal and robust cost-invalidity trade-offs when increasing training data.}
    \label{fig:data_size}
    \vspace{-5pt}
\end{figure}

\begin{figure}[ht]
    \centering
    \includegraphics[width=0.8\columnwidth]{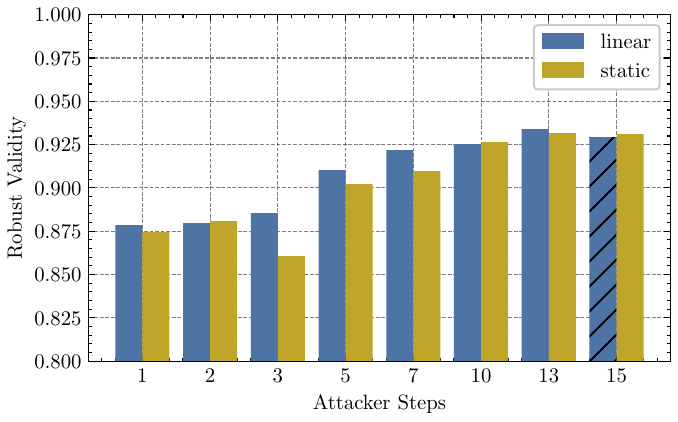}
    \caption{Impact of epsilon scheduler on the robustness. The curriculum training boosts the robustness of recourses.}
    \label{fig:eps_scheduler}
    \vspace{-5pt}
\end{figure}

%As a result, Figure~\ref{fig:n_steps_robustness} and \ref{fig:eps_robustness} further illustrate the cost-invalidity trade-off.
% when increasing the attacker's strength (by increasing the attacker steps), the robust validity increases (see Figure~\ref{fig:n_steps_rob_val}), but at the same time, the proximity also increases (see Figure~\ref{fig:n_steps_prox}). 

% Note that RoCourseNet does not achieve the best performance in terms of proximity. This is expected due to need to balance the trade-off between cost of change (e.g., proximity) and invalidation percentage (or invalidity, defined as 1 - \emph{validity}) \citep{rawal2020can}. Conceptually, to enlarge the validity of CF explanations, the cost of changes will naturally increase. Furthermore, to achieve high robust validity, we need to further sacrifice the proximity for robustness. Given the high robust validity of CF examples generated from RoCourseNet, it is natural to have a relatively large proximity value as compared to other baselines which have low robust validity.

\subsection{Further Analysis}
\label{sec:analysis}

\noindent\textbf{RoCourseNet Training Requires More Data.}
We further delve into the impact of data size on the cost-invalidity trade-off. Figure~\ref{fig:data_size} illustrates the Pareto Frontier plot of RoCourseNet trained on various fractions of the \emph{Loan} dataset. This figure shows that increasing the training data leads to a better balance between normal and robust cost-invalidity trade-offs. This supports our hypothesis that the underperformance of RoCourseNet on the \emph{German Credit} and \emph{Student} datasets can be attributed to their limited number of data points, making adversarial training more challenging.
This result also echoes findings in prior literature \citep{schmidt2018adversarially}, which shows that the sample complexity of robust learning can be significantly larger.
\\

\noindent\textbf{Evaluating VDS Attacker.}
We demonstrate the effectiveness of VDS on solving the inner maximization problem in Eq.~\ref{eq:adv_training}, i.e., how often VDS succeeds in finding an adversarial shifted model $f(.;\theta')$, such that the generated recourses $\cf$ are not robustly valid. 
To evaluate the performance of VDS, we apply Algorithm~\ref{alg:metashift} to find shifted model parameters $\theta'$, and compute the robust validity of all hold-out test-sets with respect to this shifted model.

Figure~\ref{fig:n_steps_attacker} and \ref{fig:eps_attacker} compares the effectiveness of attacking RoCourseNet and CounterNet via the VDS algorithm, which highlights three important findings: 
(i) First, the VDS algorithm is effective in finding a shifted model which invalidates a given recourse - the average robust validity drops to 74.8\% when the attacker steps $T=20$, as compared to 99.8\% validity on the original model.
(ii) Figure~\ref{fig:n_steps_attacker} shows that when $T$ is increased, the robust validity of both CounterNet and RoCourseNet is degraded, which indicates that increasing attack steps improves the effectiveness of solving the bi-level problem in Eq.~\ref{eq:bi-level}. Similarly, Figure~\ref{fig:eps_attacker} shows that increasing $E$ also improves effectiveness of solving Eq.\ref{eq:bi-level}. 
(iii) Finally, RoCourseNet is more robust than CounterNet when attacked by the VDS algorithm, as RoCourseNet vastly outperforms CounterNet in robust validity (e.g., $\sim$28\%, $\sim$10\% improved robust validity when $T=20$, $E=0.5$ in Figure~\ref{fig:n_steps_attacker} and \ref{fig:eps_attacker}, respectively). \\

% \jc{maybe change it layout to 2x3 and use wrapfig to save space?}
% 0.1 step size; RoCOurseNet T=7

\noindent\textbf{Understanding the Tri-level Robust CF Training.} 
We further analyze our tri-level robust training procedure. First, a stronger attacker (i.e., more effective in solving Eq. \ref{eq:bi-level}) leads to the training of a more robust CF generator.
In Figure~\ref{fig:n_steps_rob_val}, we observe that increasing $T$ (which results in a stronger attacker as shown in Figure~\ref{fig:n_steps_attacker}) leads to improved robust validity, which indicates a more robust CF generator. Similarly, Figure~\ref{fig:eps_rob_val} illustrates that increasing $E$ values leads to improved robust validity. 
These results show that having an appropriately strong attacker (i.e., effectively optimizing Eq.~\ref{eq:bi-level}) is crucial to train for a robust CF generator. \\

\noindent \textbf{Epsilon Scheduler.} 
Figure~\ref{fig:eps_scheduler} illustrates the importance of linearly scheduling $\epsilon$ values inside VDS. %(see Appendix for experiments with non-linear schedulings). 
By linearly increasing $\epsilon$, we observe $\sim$0.88\% improved robust validity (on average) compared to using a static $\epsilon$ during the entire adversarial training. This shows that this curriculum training strategy can boost the robustness of recourses.

% \begin{figure}
%     \centering
%     \includegraphics[width=0.5\textwidth]{figs/eps_scheduler.pdf}
%     \caption{Impact of epsilon scheduler on the robustness.}
% \end{figure}

% Fig. batchsize vs robustness

% Fig: (left) adversarial loss curve; (right) robust validity curve. This figure illustrates the overfitting problem.

% \textbf{Tri-level training suffers less in the invalidity-cost trade-off.}

% Fig: bar chart (with st.d.) comparison of different CF methods

% Fig: (left) validity vs cost; (right) robust validity vs cost

\section{Generalizing RoCourseNet to Parametric CF Explanation Methods}

We now discuss how the RoCourseNet framework is general enough to be used with other parametric CF explanation methods; in fact, we illustrate how the \emph{RoCourseNet framework can be used to improve the robustness of any parametric CF explanation method}.
%This section discusses extending the training of RoCourseNet (see Algorithm~\ref{alg:metashift} \& \ref{alg:adv_training}) into a general-purpose framework to improve the robustness of \emph{any post-hoc parametric CF explanation methods}.

Note that the VDS algorithm and its corresponding adversarial training procedure (Algorithms 1 \& 2) require access to (i) the training dataset $(x,y) \in \mathcal{D}$ and (ii) the weights of the predictive models $\theta_f$.
Therefore, any parametric CF explanation method that satisfies these assumptions can leverage these algorithms to improve their robustness to data shift. 

Thus, while in this paper, we have made a conscious decision of choosing CounterNet as the CF explanation method of choice within the RoCourseNet framework (since CounterNet's end-to-end architecture addresses the limitations of post-hoc approaches); in general, the RoCourseNet framework is model-agnostic, and it can work with any parametric model based CF explanation method.

%Importantly, this general-purpose framework is model-agnostic, as it does not require a tailor-made network structure (i.e., the CounterNet architecture).

% To use the RoCourseNet framework with any other parametric CF explanation method, we utilize a slightly altered version of Eq.~\ref{eq:adv_training} as the objective function:
We formulate a slightly altered version of Eq.~\ref{eq:adv_training} as the objective function to use the RoCourseNet framework:% with any other parametric CF explanation method:
\begin{equation}
    \begin{aligned}
    \label{eq:general}
    &\operatorname*{argmin}_{\theta, \theta_g}
    \frac{1}{N}\sum\nolimits_{(x_i,y_i) \in \mathcal{D}}\ 
    \bigg[
    \lambda_3 \cdot \underbrace{\mathcal{L}\Big(x_i, g(x_i; \theta_g)\Big)}_{\text{Proximity Loss} \ ({L}_3)} \bigg]
    \\
    &+  \max_{\boldsymbol{\delta}, \forall \delta_i \in \Delta}
    % \mathbb{E}_{\left(x_i, y\right) \sim D}
    \frac{1}{N}\sum\nolimits_{\left(x_i, y_i\right) \in \mathcal{D}}
    \bigg[ 
    \lambda_2 \cdot \underbrace{\mathcal{L}\Big(f\left(\cf_i; \theta^{\prime}_{opt}(\boldsymbol{\delta}) \right), 1 - f\left(x_i; \theta\right)\Big)}_{\text{Robust Validity Loss} \ ({L}_2)}
    \bigg] \\
    & s.t., \ {\theta'_{opt}}(\boldsymbol{\delta}) = \operatorname{argmin}_{\theta'}
        % \mathbb{E}_{(x,y)\sim \mathcal{D}}
        \frac{1}{N}\sum\nolimits_{\left(x_i, y_i\right) \in \mathcal{D}}
        \bigg[\mathcal{L}\Big(f(x_i + \delta_i; \theta'), y_i\Big)\bigg].
    \end{aligned}
\end{equation}
where $g(\cdot; \theta_g)$ represents a CF model $g: \mathcal{X} \to \mathcal{X}^\text{cf}$ parametrized by $\theta_g$. Note that Eq.~\ref{eq:general} is almost identical to Eq.~\ref{eq:adv_training}, except that the prediction loss ($\mathcal{L}_1$) is ignored, as post-hoc parametric CF explanation methods do not jointly train predictions and CF explanations (instead, they assume access to a pre-trained ML model).

To solve Eq.~\ref{eq:general}, we can apply a similar strategy used while training RoCourseNet: for each mini-batch datapoints $\{x^{(i)}, y^{(i)} \}^m$, (i) we solve the inner bi-level problem to obtain the worst-shift weight of the predictive model $\theta_f'$ by applying the VDS algorithm (outlined in Algorthim~\ref{alg:metashift}); (ii) next, we solve the outer minimization problem by optimizing against the shifted predictive model. \\

% How to do that
% Formally, to robustly train the weight $\theta_g$ of a CF explanation, we first 
% \\

% Baselines
\noindent\textbf{Experiment Setting.} To demonstrate the generalizability of the RoCourseNet framework, we experiment with two post-hoc parametric CF explanation methods and their ``\textit{robustified}" models.
\begin{itemize}[leftmargin=*]
    \item \textsc{CF Model} is a multi-layer perceptron model which outputs recourses $\cf$ given $x$. This model is optimized for validity and proximity. We also train a \textsc{Robust CF Model} to generate robust recourses using the RoCourseNet framework with \textsc{CF Model}. %switches the validity loss into the robust validity loss (in Eq.~\ref{eq:bi-level}). 
    \item \textsc{VAE-CF} is a well-known parametric method that uses variational auto-encoder (VAE) to generate recourses $\cf$ given input $x$. Similarly, \textsc{Robust VAE-CF} utilizes the RoCourseNet framework to generate robust recourses with \textsc{CF Model}.
\end{itemize}
For fair comparison, the architecture of \textsc{CF Model} is the same as the network that combines the encoder and CF generator.% (as shown in Figure \ref{fig:architecture}). 
Our experiment is evaluated on the \emph{Loan} dataset, which follows the same experiment settings in Section~\ref{sec:experiment} (e.g., data partitioning, metrics, etc).
\\

\begin{table}[t]
\centering
\small
\caption{\label{tab:framework}Evaluating robustness under model shift with generalized parametric models on the \emph{Adult} dataset. The \textsc{robust} training \textcolor{blue}{improves} both validity and  robust validity.}
\begin{tabular}{@{}l|cccc@{}}
\toprule
\textbf{Methods}                          & \multicolumn{3}{c}{\textbf{Metrics}}                                                               \\ 
% \cmidrule(lr){2-4}
\multicolumn{1}{c|}{\textbf{}}    
& \cellcolor{cid} {Prox.} & \cellcolor{cood} {Val.} &  \cellcolor{ccon} {Rob-Val.}
\\ \midrule\midrule
% \multirow{3}{*}{\textbf{Loan}}         
\textsc{CF Model} & 5.753 $\pm$ 0.647  & 0.826 $\pm$ 0.1825 &  0.607 $\pm$ 0.167  \\
\textsc{Robust CF Model} & 0.739 $\pm$ 1.078  & \textcolor{blue}{0.949 $\pm$ 0.022} &  \textcolor{blue}{0.906 $\pm$ 0.119}  \\ \midrule 
\textsc{VAE-CF} & 8.531 $\pm$ 1.241 &  0.745 $\pm$ 0.272 & 0.692 $\pm$ 0.251  \\
\textsc{Robust VAE-CF} &  9.473 $\pm$ 0.962 & \textcolor{blue}{0.818 $\pm$ 0.186} & \textcolor{blue}{0.807 $\pm$ 0.189} \\ \midrule
% \textsc{Roar-LIME} & & 7.648 $\pm$ 2.248 &   0.937 $\pm$ 0.046   &   0.908 $\pm$ 0.107 \\
\rowcolor[HTML]{EFEFEF}
\textsc{RoCourseNet} &  {6.611 $\pm$ 0.418} &  \textbf{0.996 $\pm$ 0.002} &    \textbf{0.969 $\pm$ 0.106}  \\\bottomrule
\end{tabular}
\end{table}

\begin{figure}[t]
     \centering
     \begin{subfigure}[h]{\columnwidth}
         \centering
         \includegraphics[width=0.93\textwidth]{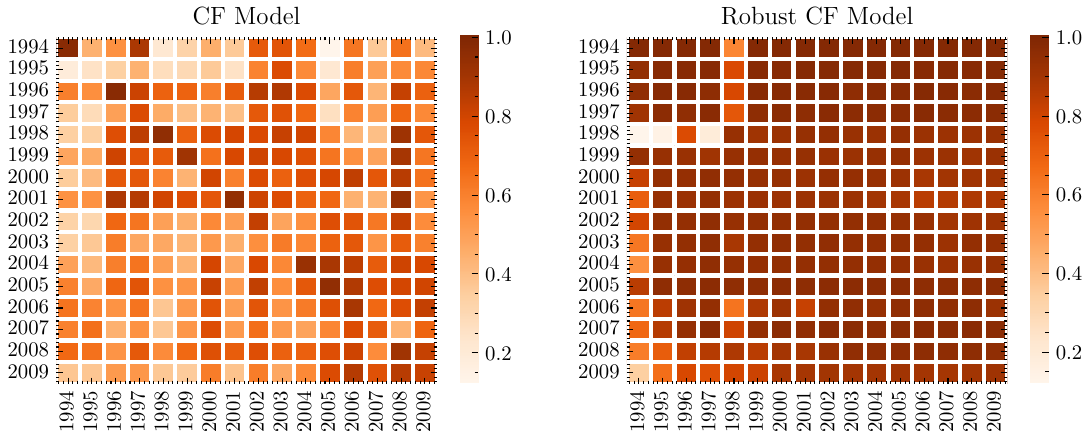}
         \caption{Robust validity matrix of CF Model \& Robust CF Model.}
     \end{subfigure}
     \hfill
     \begin{subfigure}[h]{\columnwidth}
         \centering
         \includegraphics[width=0.93\textwidth]{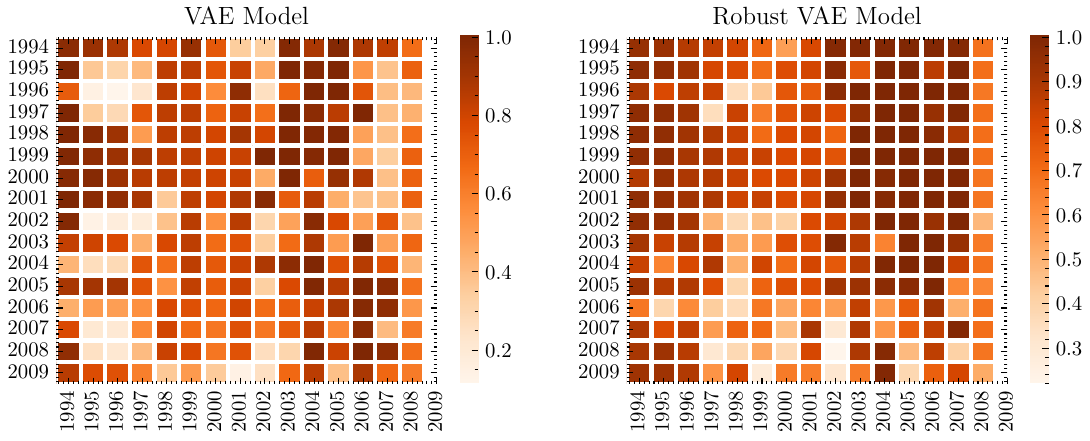}
         \caption{Robust validity matrix of VAE-CF \& robust VAE-CF.}
     \end{subfigure}
     % \hfill
     % \begin{subfigure}[h]{\columnwidth}
     %     \centering
     %     \includegraphics[width=0.93\textwidth]{figs/validity_matrix_cfnet.pdf}
     %     \caption{The validity matrix of (left) CounterNet \& (right) RoCourseNet.}
     % \end{subfigure}
    \caption{Comparing the robust validity matrix between (left) normal and (right) adversarial training on two parametric CF methods. Darker color indicates higher robust validity. The proposed adversarial training improves recourse robustness.}
    \label{fig:val_matrix}
    \vspace{-5pt}
\end{figure}

\noindent\textbf{Empirical Results.}
Table~\ref{tab:framework} compares the (robust) \textsc{CF Model} and \textsc{VAE-CF} with RoCourseNet (the best performing method). The results demonstrate two key findings:
(i) First, our proposed tri-level robust training (in Algorithm~\ref{alg:adv_training}) is general purpose and can be plugged in as-is to improve the robustness of post-hoc parametric CF methods. 
In particular, applying robust training to both \textsc{CF Model} and \textsc{VAE-CF} improves the robust validity of these models by $\sim 32.5\%$ and 17.3\% (on average), respectively. 
Figure~\ref{fig:val_matrix} further illustrates this finding, where robustly trained methods (via the generalized RoCourseNet framework) generate recourses with higher robust validity (i.e., the robust validity matrix has more darker colored elements).
(ii) Additionally, 
the design of RoCourseNet, utilizing the design of CounterNet \citep{guo2021counternet}, proves to be effective in balancing the cost-invalidity trade-off, as we observe that RoCourseNet outperforms \textsc{Robust CF Model} and \textsc{Robust VAE-CF} in proximity, validity, and robust validity.
The result shows the advantages of RoCourseNet's joint training of prediction and robust recourses.

% we highlight the importance of RoCourseNet design (by leveraging the design of CounterNet \citep{guo2021counternet}). This architecture design enables to generation of well-aligned CF explanations by passing the predictor model's decision boundary (i.e., $p_x$) to the CF generator. From Table~\ref{tab:framework}, we observe that RoCourseNet outperforms this Robust CF Framework in terms of the validity and robust validity, which underscores the importance of RoCourseNet's architecture designs.

\section{Conclusion}
\label{sec:conclusion}

We present \emph{RoCourseNet}, an end-to-end training framework to generate predictions and robust CF explanations. We formulate this robust end-to-end training as a tri-level optimization problem, and leverage novel adversarial training techniques to solve this problem. Empirical results show that RoCourseNet outperforms state-of-the-art baselines in robust validity, and achieves better balance on the cost-validity trade-off. We further demonstrate that the RoCourseNet training framework is generalizable to be applied with any parametric CF explanation method.
\bibliographystyle{ACM-Reference-Format}
\balance
\bibliography{ref}

\newpage
\appendix
\section{Implementation Details}
\label{sec:implementation}
\label{sec:app-implementation}
Here we provide implementation details of RoCourseNet and three baseline methods on three datasets listed in Section~\ref{sec:experiment}. The code can be found through this anonymous repository (\url{https://www.dropbox.com/s/gsrpt55hf2ik7v4/RoCourseNet.zip?dl=0}).

% \begin{figure}[t!]
% \centering
% \includegraphics[width=0.9\linewidth]{figs/architecture-1.pdf}
% \caption{The architecture of CounterNet, the building block of RoCourseNet \citep{guo2021counternet}.}
% \label{fig:architecture}
% \end{figure}

\textbf{Feature Engineering.}
We follow the feature engineering procedure of CounterNet \citep{guo2021counternet}.
Specifically, for continuous features, we scale all feature values into the $[0,1]$ range.
To handle the categorical features, we customize model architecture for each dataset. First, we transform the categorical features into numerical representations via one-hot encoding. 
In addition, for each categorical feature, we add a softmax layer after the final output layer in the CF generator, which ensures that the generated CF examples respect the one-hot encoding format.

\textbf{Hyperparameters.}
For all three datasets, we train the model for up to 50 epochs with Adam. We set dropout rate to 0.3 to prevent overfitting. We use $T=7$ and $E=0.1$ to report results in Table~\ref{tab:cf}, and report the impact of attacker steps $T$ and maximum perturbation $E$ to robustness in Figure~\ref{fig:robustness}. 
We use $K=2$ unrolling steps (same as \cite{huang2020metapoison}) with the step size $\alpha = 2.5 \times \delta / T$ (based on \cite{madry2017towards}) for solving the bi-level problem in Equation~\ref{eq:bi-level} (via VDS).
In addition, Table~\ref{tab:param} reports the hyperparameters chosen for each dataset, and Table~\ref{tab:arch} speficies the architecture used for each dataset.

\begin{table*}[htp]
%\vskip 0.1in
\centering
\caption{\label{tab:param}Hyperparameters setting for each dataset.}
\begin{tabular}{lcccccc}
\toprule
\textbf{Dataset} &
  \multicolumn{1}{l}{\textbf{\begin{tabular}[c]{@{}l@{}}Learning Rate\end{tabular}}} &
  $\boldsymbol{\eta}$ &
  \multicolumn{1}{l}{\textbf{\begin{tabular}[c]{@{}l@{}}Batch Size\end{tabular}}} &
  $\boldsymbol{\lambda_1}$ & $\boldsymbol{\lambda_2}$ & $\boldsymbol{\lambda_3}$
  \\ 
  \midrule
  \textbf{Loan} & 0.003 & 0.03 & 128 & 1.0 & 0.2 & 0.1 \\
  \textbf{German Credit}  &  0.003 & 0.03 & 256  & 1.0 & 1.0 & 0.1 \\
  \textbf{Student} & 0.01 & 0.01 & 128 & 1.0 & 0.2 & 0.1  \\\bottomrule
\end{tabular}
\end{table*}

\begin{table*}[htp]
%\vskip 0.1in
\centering
\caption{\label{tab:arch}Architecture specification of RoCourseNet for each dataset.}
\begin{tabular}{lccccc}
\toprule
\textbf{Dataset} &
  \multicolumn{1}{l}{\textbf{\begin{tabular}[c]{@{}l@{}}Encoder Dims\end{tabular}}} &
  \multicolumn{1}{l}{\textbf{\begin{tabular}[c]{@{}l@{}}Predictor Dims\end{tabular}}} &
  \multicolumn{1}{l}{\textbf{\begin{tabular}[c]{@{}l@{}}CF Generator Dims\end{tabular}}}\\ 
  \midrule
  \textbf{Loan}  & [110,200,10] & [10, 10] & [10, 10] \\
  \textbf{German Credit} & [19, 100,10] & [10, 20] & [10, 20] \\
  \textbf{Student}  & [83,50,10] & [10, 10] & [10, 50] \\\bottomrule
\end{tabular}
\end{table*}

\textbf{Software and Hardware Specifications.} We use Python (v3.7) with Pytorch (v1.82), Pytorch Lightning (v1.10), numpy (v1.19.3), pandas (v1.1.1), scikit-learn (v0.23.2) and higher (v0.2.1) \cite{grefenstette2019generalized} for the implementations. All our experiments were run on a Debian-10 Linux-based Deep Learning Image on the Google Cloud Platform. The RoCourseNet and baseline methods are trained (or optimized) on a 16-core Intel machine with 64 GB of RAM.

\section{Additional Experimental Analysis}

\subsection{Predictive Performance}

We first show that, similar to CounterNet, the training of RoCourseNet does not come at the cost of degraded predictive accuracy. Table~\ref{tab:acc} \& \ref{tab:auc} compare RoCourseNet's predictive accuracy and AUC score against the base prediction model used by baselines. This table shows that RoCourseNet achieves competitive predictive performance -- it achieves marginally better accuracy than the base model ($\sim2\%$). Thus, we conclude that the joint training of RoCourseNet does not come at a cost of reduced predictive performance.

% \textbf{Predictive Accuracy.} Table \ref{tab:accuracy} compares CounterNet's predictive accuracy against the base prediction model used by baselines. This table shows that CounterNet exhibits highly competitive predictive performance - it achieves marginally better accuracy on the Student and Titanic datasets (row 2 \& 3), and achieves marginally lower accuracy on the remaining datasets. Across all six datasets, the difference between the predictive accuracy of CounterNet and the base model is $\sim$ 0.1\%. Thus, the potential benefits achieved by CounterNet's joint training of predictor and CF generator networks do not come at a cost of reduced predictive accuracy.%, as CounterNet achieves highly competitive performance against an ML model solely optimized for predictive accuracy.

\begin{table}[htp]
%\vskip 0.1in
\centering
\caption{\label{tab:acc}Predictive accuracy for each dataset.}
\begin{tabular}{lccc}
\toprule
\textbf{Dataset} &
  \multicolumn{1}{l}{\textbf{\begin{tabular}[c]{@{}l@{}}Base Model\end{tabular}}} &
  \multicolumn{1}{l}{\textbf{\begin{tabular}[c]{@{}l@{}}RoCourseNet\end{tabular}}} \\ 
  \midrule
  \textbf{Loan} & 0.886 $\pm$ 0.036 & 0.885 $\pm$ 0.035 \\
  \textbf{German Credit}  &  0.714 $\pm$ 0.003 & 0.742 $\pm$ 0.014  \\
  \textbf{Student} & 0.914 $\pm$ 0.028 & 0.906 $\pm$ 0.066 \\\bottomrule
\end{tabular}
\end{table}

\begin{table}[htp]
%\vskip 0.1in
\centering
\caption{\label{tab:auc}AUC score for each dataset.}
\begin{tabular}{lccc}
\toprule
\textbf{Dataset} &
  \multicolumn{1}{l}{\textbf{\begin{tabular}[c]{@{}l@{}}Base Model\end{tabular}}} &
  \multicolumn{1}{l}{\textbf{\begin{tabular}[c]{@{}l@{}}RoCourseNet\end{tabular}}} \\ 
  \midrule
  \textbf{Loan} & 0.897 $\pm$ 0.026 & 0.900 $\pm$ 0.027 \\
  \textbf{German Credit}  &  0.662 $\pm$ 0.018 & 0.729 $\pm$ 0.010 \\
  \textbf{Student} & 0.913 $\pm$ 0.012 & 0.947 $\pm$ 0.018 \\\bottomrule
\end{tabular}
\end{table}

\subsection{Heuristic Baselines}

We provide two heuristic baseline methods to further illustrate the challenge of generating robust recourses under the distribution shift scenarios. \emph{VanillaCF-Random} aims to generate robust recourse by adding a small perturbation to input. In addition, \emph{RoCourseNet-Random} optimizes for robust CF generator against a random perturbation attacker.

Table~\ref{tab:random} compares heuristic baselines with RoCourseNet. Both baseline methods peroform significantly worse than RoCourseNet in validity, robust validity and proximity. This experiment further highlights the hardness of generating robust recourses as simple heuristics drastically underperform as compared to RoCourseNet.

\begin{table*}[htp]
%\vskip 0.1in
\centering
\caption{\label{tab:random}Heuristic baselines as compared to RoCourseNet on Loan dataset. Simple heuristic does not defend against distribution shift.}
\begin{tabular}{l|ccc}
\toprule
\textbf{Method} & \textbf{Validity} & \textbf{Robust Validity} & \textbf{Proximity}\\ 
  \midrule
  \textsc{VanillaCF-Random} & 0.634$\pm$0.270  & 0.510$\pm$0.251 & 9.446$\pm$1.042 \\
  \textsc{RoCourseNet-Random}  &  0.856 $\pm$0.127 & 0.856 $\pm$0.127 & 10.405 $\pm$ 1.857 \\
  \textsc{RoCourseNet} & \textbf{0.996 $\pm$ 0.002} & \textbf{0.969 $\pm$ 0.106} & \textbf{6.611 $\pm$ 0.418} \\\bottomrule
\end{tabular}
\end{table*}

\subsection{Simulated Data Shift}

We conduct simulated experiments with covariant and label shift. 

\begin{figure*}
    \centering
    \includegraphics[width=\linewidth]{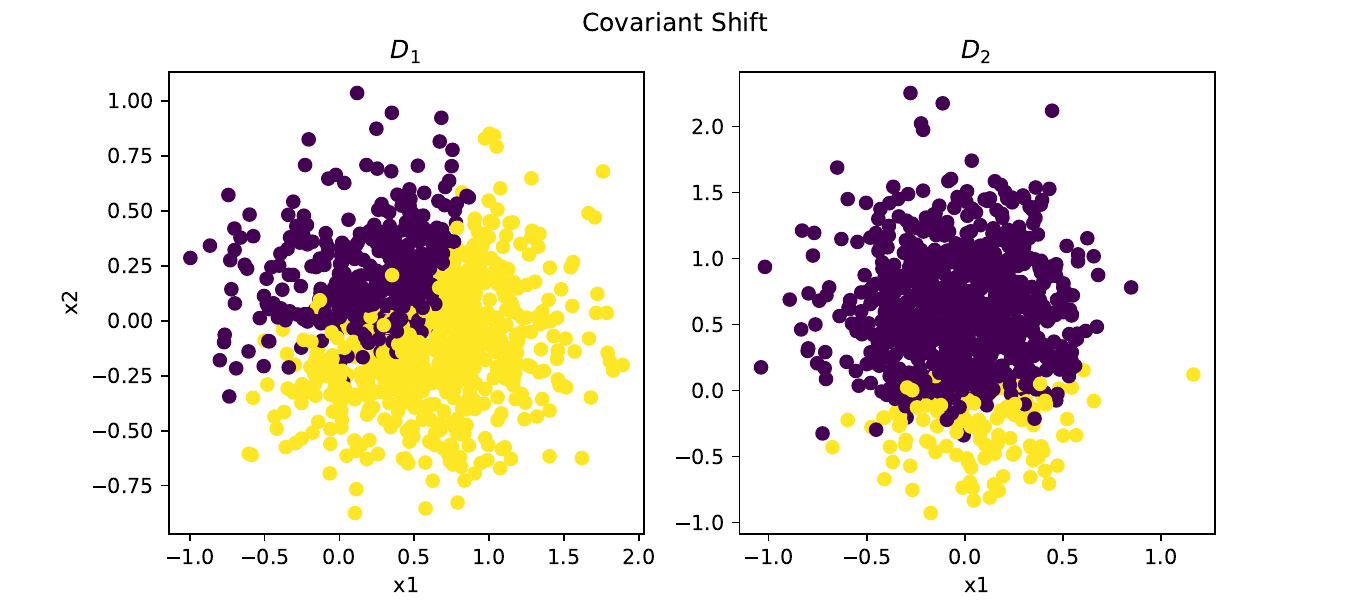}
    \caption{Illustration of covariant shift.}
    \label{fig:cov_shift}
\end{figure*}

\paragraph{Covariant Shift}
We simulate the covariant shift via this Bayesian network:

\begin{align*}
    x_1 & \sim \mathcal{N}(\mu_1,\sigma_1) \\
    x_2 & \sim \mathcal{N}(\mu_2,\sigma_2) \\
    y & = -x_2 + x_1^3 + \varepsilon,\ \varepsilon \sim \mathcal{N}(-0.1, 0.1)
\end{align*}
where we set $\mu_1=0.5,\sigma_1=0.5, \mu_2=0,\sigma_2=0.3$ for $D_1$, and $\mu_1=0,\sigma_1=0.3,\mu_2=0.5,\sigma_2=0.5$ for $D_2$. Figure~\ref{fig:cov_shift} illustrates this simulation dataset.

\paragraph{Label Shift}

\begin{figure*}
    \centering
    \includegraphics[width=\linewidth]{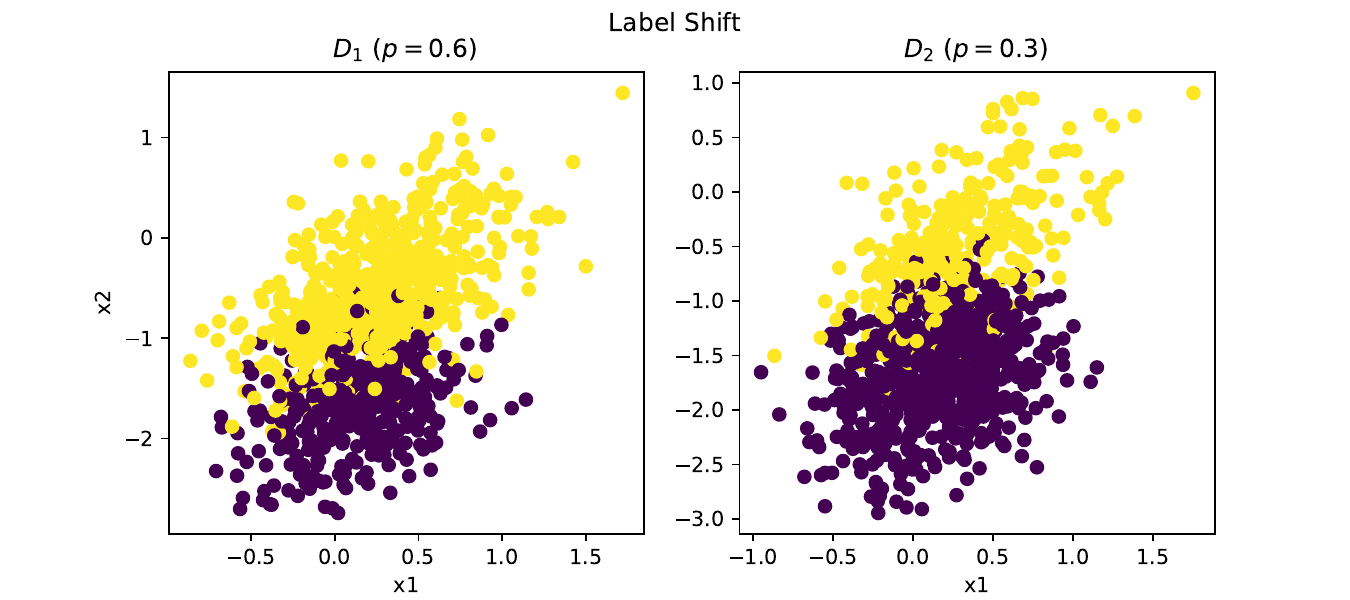}
    \caption{Illustration of label shift.}
    \label{fig:label_shift}
\end{figure*}

Similarly, we simulate the label shift via this Bayesian network:
\begin{align*}
    y & \sim binomial(p) \\
    z & = 2y - 1 + \varepsilon,\ \varepsilon \sim \mathcal{N}(0.1, 0.1) \\
    x1 & \sim  \mathcal{N}(-z + z^3, 0.3) \\
    x2 & \sim  \mathcal{N}(z + z^3 - 3y , 0.3) \\
\end{align*}
where we set $p=0.6$ for $D_1$, and $p=0.3$ for $D_2$. Figure~\ref{fig:label_shift} illustrates this simulation dataset.

\paragraph{Experimental Results}
Table~\ref{tab:shifts} shows that RoCourseNet achieves 100\% validity and robust validity under both covariate and label shifts.

\begin{figure}
    \centering
    \includegraphics[width=\linewidth]{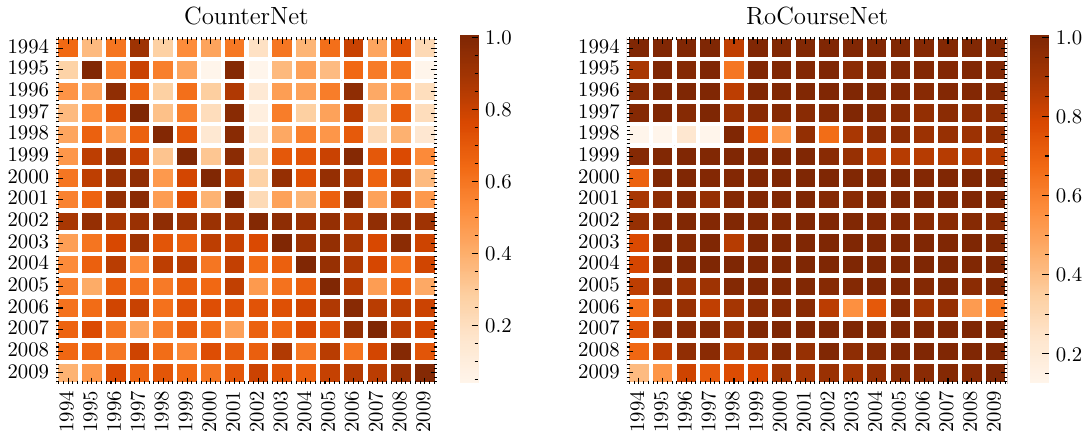}
    \caption{Validity matrix of CounterNet and RoCourseNet. RoCourseNet significantly improves the recourse robustness.}
    \label{fig:val_matrix_cfnet}
\end{figure}

\paragraph{Validity Matrix of CounterNet and RoCourseNet.}
Figure~\ref{fig:val_matrix_cfnet} shows the validity matrix of CounterNet and RoCourseNet.

\begin{table*}[htp]
%\vskip 0.1in
\centering
\caption{\label{tab:shifts}RoCourseNet on simulated data shifts.}
\begin{tabular}{lccc}
\toprule
\textbf{Data Shift} & 
\textbf{Validity} &
\textbf{Robust Validity} &
\textbf{Proximity} \\ 
  \midrule
  \textbf{Covariant} & 1.00 & 1.00 & 0.389 $\pm$ 0.051 \\
  \textbf{Label} & 1.00 & 1.00 & 0.425 $\pm$ 0.012 \\\bottomrule
\end{tabular}
\end{table*}

\section{Ablations of RoCourseNet}

\subsection{Training Loss Curve of RoCourseNet}

\begin{figure}
    \centering
    \begin{subfigure}[h]{0.3\textwidth}
         \centering
         \includegraphics[width=0.93\textwidth]{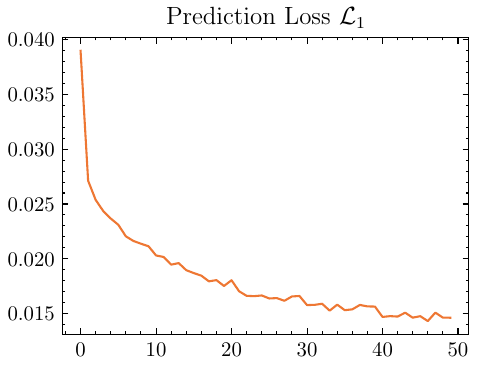}
         \caption{$\mathcal{L}_1$ training curve.}
         \label{fig:adult_ablation}
    \end{subfigure}
    \hfill
    \begin{subfigure}[h]{0.3\textwidth}
         \centering
         \includegraphics[width=0.93\textwidth]{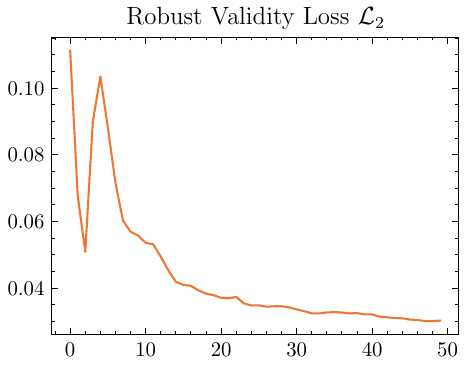}
         \caption{$\mathcal{L}_2$ training curve.}
         \label{fig:home_ablation}
    \end{subfigure}
    \hfill
    \begin{subfigure}[h]{0.3\textwidth}
         \centering
         \includegraphics[width=0.93\textwidth]{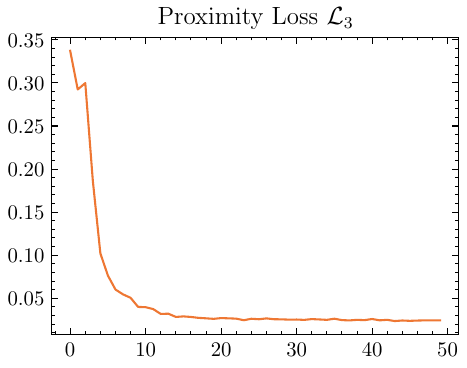}
         \caption{$\mathcal{L}_3$ training curve.}
         \label{fig:student_ablation}
    \end{subfigure}
    \caption{Training loss curves of RoCourseNet on the Loan dataset.}
    \label{fig:train_loss}
\end{figure}

Figure~\ref{fig:train_loss} shows RoCourseNet's training curve. Importantly, the prediction loss $\mathcal{L}_1$ and the proximity loss $\mathcal{L}_3$ are smoothly optimized during the training. The robust validity loss $\mathcal{L}_2$ encounters fluctuations in the early stage of training, but starts to converge after 10 epochs. 

\subsection{$l_2$-norm Projection in Algorithm~\ref{alg:metashift}}

\begin{figure}[t]
     \centering
     \begin{subfigure}[h]{0.3\textwidth}
         \centering
         \includegraphics[width=0.93\textwidth]{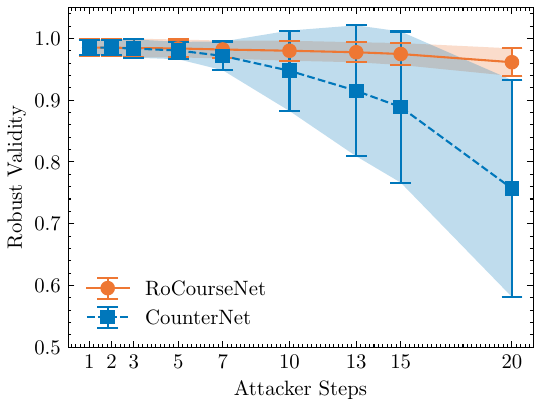}
         \caption{The number of attacker steps $T$ vs attacker effectiveness ($\downarrow$)}
     \end{subfigure}
     \hfill
     \begin{subfigure}[h]{0.3\textwidth}
         \centering
         \includegraphics[width=0.93\textwidth]{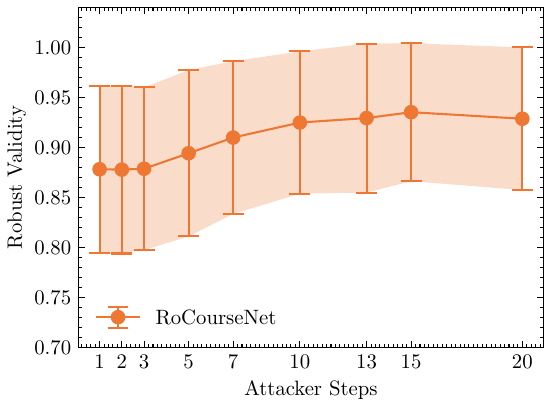}
         \caption{The number of attacker steps $T$ vs robust validity ($\uparrow$).}
     \end{subfigure}
     \hfill
     \begin{subfigure}[h]{0.3\textwidth}
         \centering
         \includegraphics[width=0.93\textwidth]{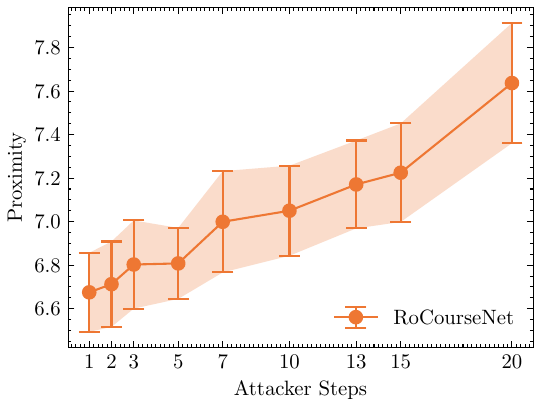}
         \caption{The number of attacker steps $T$ vs proximity ($\downarrow$).}
     \end{subfigure}
    \caption{The impact of the number of attacker steps $T$ under the $l_2$-norm constrains.}
    \label{fig:robustness_l2}
\end{figure}

We provide supplementary results on adopting $\Delta$ as the $l_2$-norm ball (i.e., $\Delta = \{\delta \in \mathbb{R}^n\ |\ ||\delta||_2 \leq \epsilon \}$) for the maximum perturbation constrains. Figure~\ref{fig:robustness_l2} highlights the results of using the $l_2$-norm ball in attacking and adversarial training. We observe similar patterns in Figure~\ref{fig:robustness}.
Thus, this result shows that $l_\infty$-norm constrain can be substitute to other feasible region.
 
% to project  analysis of other two large datasets (HELOC and OULAD, shown in Figure~\ref{fig:home_ablation} \& \ref{fig:student_ablation}, respectively). We observe the similar pattern as Figure~\ref{fig:ablation} in the original paper.

\section{Time-Complexity Analysis}

\begin{table}[tp]
\small
\centering
\small
\footnotesize
\caption{\label{tab:train_time}Training time of CounterNet and RoCourseNet on the Loan dataset (the largest dataset).}
\begin{tabular}{l|ccc}
\toprule
\textbf{Dataset} &
  \multicolumn{1}{l}{\textbf{\begin{tabular}[c]{@{}l@{}}CounterNet\end{tabular}}} &
  \multicolumn{1}{l}{\textbf{\begin{tabular}[c]{@{}l@{}}RoCourseNet\end{tabular}}} \\ 
  \midrule
  \textbf{Loan} & 22m 6s & 32m 16s \\
  \textbf{German Credit} & 46s  & 1m 37s \\
  \textbf{Student} & 55s & 2m 19s \\
  \bottomrule
\end{tabular}
\end{table}

\subsection{Training Time}
RoCourseNet takes only $\sim$10 more minutes of training (as compared to CounterNet) on the Loan dataset (our largest-sized dataset). This is quite reasonable since training time is a one-time up-front cost; after RoCourseNet is trained, test-time inference happens in milliseconds. Table~\ref{tab:train_time} shows the training time of these two models.

\begin{table}[tp]
\small
\centering
\small
\footnotesize
\caption{\label{tab:runtime}Inference time for generating a single CF example on the Loan dataset (in milliseconds).}
\begin{tabular}{l|cccc}
\toprule
\textbf{Dataset} &
  \multicolumn{1}{l}{\textbf{\begin{tabular}[c]{@{}l@{}}ROAR\end{tabular}}} &
  \multicolumn{1}{l}{\textbf{\begin{tabular}[c]{@{}l@{}}RBR\end{tabular}}} &
  \multicolumn{1}{l}{\textbf{\begin{tabular}[c]{@{}l@{}}CounterNet\end{tabular}}} &
  \multicolumn{1}{l}{\textbf{\begin{tabular}[c]{@{}l@{}}RoCourseNet\end{tabular}}} \\ 
  \midrule
  \textbf{Loan} & 131.38 & 345.91 & 0.67 & 0.67 \\
  \textbf{German Credit} & 104.87 & 271.34 & 0.51 & 0.51 \\
  \textbf{Student} & 213.58 & 555.91 & 1.00 & 1.01 \\
  
  \bottomrule
\end{tabular}
\end{table}

\subsection{Inference Time}

Inference runtime is an important metric as recourses are user-facing. Table~\ref{tab:runtime} shows the inference runtime of RoCourseNet and baseline methods. Importantly, CounterNet and RoCourseNet achieve the same amount of inference time (as they share the same network structure). On the other hand, ROAR and RBR (two non-parametric methods) take significantly more time (i.e., $\sim$200X and $\sim$500X runtime as compared to RoCourseNet, respectively).

\section{Discussion about Multi-class Classification}

Existing CF explanation literature focuses on evaluating methods under the binary classification settings \citep{mothilal2020explaining, mahajan2019preserving, upadhyay2021towards, guo2021counternet}. However, these CF explanation methods can be adapted to the multi-class classification settings. Given an input instance $x\in \mathbb{R}^d$, the RoCourseNet generates  (i) a prediction $\hat{y}_x \in \mathbb{R}^k$ for input instance $x$, and (ii) a CF example $\cf$ as an explanation for input instance $x$. The prediction $\hat{y}_x \in \mathbb{R}^k$ is encoded as one-hot format as $\hat{y}_x \in \{0,1\}^k$, where $\sum_i^k \hat{y}_x^{(i)} = 1$, $k$ denotes the number of classes. In addition, we assume a desired outcome $y^\prime$ for every input instances $x$. As such, we can adapt Eq.~\ref{eq:adv_training} for binary settings to the multi-class settings as follows:
% \begin{equation}
%     \begin{aligned}
%     \label{eq:multiclass}
%     \operatorname*{argmin}_{\mathbf{\theta}=\{\theta_h, \theta_m, \theta_g\}}
%     \mathbb{E}_{\left(x, y\right) \sim D}
%     % \frac{1}{N}\sum
%     \bigg[ 
%     &\lambda_1 \cdot \underbrace{\mathcal{L}\Big(f(x; \theta_f), y\Big)}_{\text{Prediction Loss}\ ({L}_1)}+ 
%     \lambda_3 \cdot \underbrace{\mathcal{L}\Big(x, \cf\Big)}_{\text{Proximity Loss} \ ({L}_3)} \bigg] \\
%     &+  \max_{\boldsymbol{\delta}, \forall \delta \in \Delta}
%     \mathbb{E}_{\left(x, y\right) \sim D}
%     % \frac{1}{N}\sum
%     \bigg[ 
%     \lambda_2 \cdot \underbrace{\mathcal{L}\Big(f\left(\cf; \theta^{\prime}_f(\boldsymbol{\delta})^{*} \right), y^\prime\Big)}_{\text{Robust Validity Loss} \ ({L}_2)}
%     \bigg] \\
%     %& s.t., \mathcal{F} = \{\theta'_f \ | \  \operatorname{argmin}_{\theta'_f} \mathbb{E}_{(x,y)\in D}\left[ \mathcal{L}(f(x + \delta; \theta'_f), y)\right], \ \forall \ \delta \in \mathbb{B}(0, \epsilon)\}
%     & s.t\ \ {\theta'_f}(\boldsymbol{\delta})^{*} = \operatorname{argmin}_{\theta'_f}
%         \mathbb{E}_{(x,y)\sim \mathcal{D}}\bigg[\mathcal{L}\Big(f(x + \delta; \theta'_f), y\Big)\bigg].
%     \end{aligned}
% \end{equation}

\begin{equation}
    \begin{aligned}
    \label{eq:multiclass}
    % \operatorname*{argmin}_{\mathbf{\theta}=\{\theta_h, \theta_m, \theta_g\}}
    \operatorname*{argmin}_{\theta, \theta_g}
    % \mathbb{E}_{\left(x, y\right) \sim D}
    &\frac{1}{N}\sum\nolimits_{(x_i,y_i) \in \mathcal{D}}
    \bigg[ 
    \lambda_1 \cdot \underbrace{\mathcal{L}\Big(f(x_i; \theta), y_i\Big)}_{\text{Prediction Loss}\ ({L}_1)}+ 
    \lambda_3 \cdot \underbrace{\mathcal{L}\Big(x_i, \cf_i\Big)}_{\text{Proximity Loss} \ ({L}_3)} \bigg] \\
    &+  \max_{\boldsymbol{\delta}, \forall \delta_i \in \Delta}
    % \mathbb{E}_{\left(x_i, y\right) \sim D}
    \frac{1}{N}\sum\nolimits_{\left(x_i, y_i\right) \in \mathcal{D}}
    \bigg[ 
    \lambda_2 \cdot \underbrace{\mathcal{L}\Big(f\left(\cf_i; \theta^{\prime}_{opt}(\boldsymbol{\delta}) \right), y^\prime\Big)}_{\text{Robust Validity Loss} \ ({L}_2)}
    \bigg] \\
    %& s.t., \mathcal{F} = \{\theta'_f \ | \  \operatorname{argmin}_{\theta'_f} \mathbb{E}_{(x,y)\in D}\left[ \mathcal{L}(f(x + \delta; \theta'_f), y)\right], \ \forall \ \delta \in \mathbb{B}(0, \epsilon)\}
    & s.t\ \ {\theta'_{opt}}(\boldsymbol{\delta}) = \operatorname{argmin}_{\theta'}
        % \mathbb{E}_{(x,y)\sim \mathcal{D}}
        \frac{1}{N}\sum\nolimits_{\left(x_i, y_i\right) \in \mathcal{D}}
        \bigg[\mathcal{L}\Big(f(x_i + \delta_i; \theta'), y_i\Big)\bigg],
    \\ &\cf_i = g(x_i; \theta_g).
    \end{aligned}
\end{equation}

To optimize for Eq.~\ref{eq:multiclass}, we can follow the same procedure outlined in Algorithm~\ref{alg:adv_training}. For each sampled batch, we first optimize for the predictive accuracy $\theta' = \theta - \nabla_\theta (\lambda_1 \cdot {L}_1)$. Next, we use the VDS algorithm to optimize for the inner max-min bi-level problem (in Algorithm~\ref{alg:metashift}). Finally, we optimize for the CF explanations by updating the model's weight as $\theta''_g = \theta''_g - \nabla_{\theta'_g} (\lambda_2 \cdot {L}_2 + \lambda_3 \cdot {L}_3)$.

\appendix

\end{document}